\documentclass{article}
\usepackage{theapa, rawfonts}
 
\usepackage{graphicx}
\usepackage{textcomp}
\usepackage{xcolor}
\usepackage{enumerate}   
\usepackage{caption}
\usepackage{subcaption}
\usepackage{arydshln}
\usepackage{dsfont}

\usepackage{enumitem}
\usepackage{amsthm}
\usepackage{bigints}
\usepackage{dsfont}
\usepackage{bbm}
\usepackage{bm}
\usepackage{booktabs}
\usepackage{url}
\usepackage{algorithm}
\usepackage{algorithmicx}  
\usepackage{algpseudocode}
\usepackage{textcomp}
\usepackage{dsfont}
\usepackage{multirow}
\usepackage{comment}
\usepackage{mathrsfs}
\usepackage{dblfloatfix}
\usepackage{multicol}
\usepackage{arydshln}
\usepackage{amsmath}
\usepackage{amssymb}
\usepackage[margin=3cm]{geometry}

\newtheorem{definition}{Definition}

\newcommand{\specialcell}[2][c]{%
  \begin{tabular}[#1]{@{}c@{}}#2\end{tabular}}

  \title{Detecting Change Intervals\\ with Isolation Distributional Kernel}

 \author{Yang Cao \\ charles.cao@ieee.org \\  Deakin University
   \and Ye Zhu  \\ ye.zhu@ieee.org\\  Deakin University
   \and Kai Ming Ting \\ tingkm@nju.edu.cn \\ Nanjing University 
   \and Flora D. Salim \\ flora.salim@unsw.edu.au \\ University of New South Wales
   \and Hong Xian Li \\ hong.li@deakin.edu.au \\ Deakin University 
   \and Luxing Yang  \\ y.luxing@deakin.edu.au \\ Deakin University 
   \and Gang Li \\ gang.li@deakin.edu.au \\ Deakin University}

\begin{document}

 \maketitle

\begin{abstract}
Detecting abrupt changes in data distribution is one of the most significant tasks in streaming data analysis. Although many unsupervised Change-Point Detection (CPD) methods have been proposed recently to identify those changes, they still suffer from missing subtle changes, poor scalability, or/and sensitivity to outliers. To meet these challenges, we are the first to generalise the CPD problem as a special case of the Change-Interval Detection (CID) problem. Then we propose a CID method, named \texttt{iCID}, based on a recent Isolation Distributional Kernel (IDK). \texttt{iCID} identifies the change interval if there is a high dissimilarity score between two non-homogeneous temporal adjacent intervals. The data-dependent property and finite feature map of IDK enabled \texttt{iCID} to efficiently identify various types of change-points in data streams with the tolerance of outliers. Moreover, the proposed online and offline versions of \texttt{iCID} have the ability to optimise key parameter settings. The effectiveness and efficiency of \texttt{iCID} have been systematically verified on both synthetic and real-world datasets. 
\end{abstract}

\section{Introduction}~\label{intro}

A common task in streaming data is the identification and analysis of change-points that describe events of behaviour change. 
In statistics, a change-point is a time point when an observed variable changes its behaviour in a series or process~\shortcite{basseville1993detection}.
\emph{Change-Point Detection} (CPD) is an analytical method for identifying behavioural mutations in a temporal interval of observations~\shortcite{van2020evaluation}.
It has attracted significant attention from data analysts 
because many time-critical applications demand these change-points to be examined as soon as possible, e.g., system failures and abnormal network status. CPD methods have been widely used to analyse events or time-dependent anomalies in areas such as signal processing~\shortcite{truong2020selective}, 
network traffic analysis~\shortcite{lung2012distributed} and human activity recognition~\shortcite{chamroukhi2013joint}.

Although many unsupervised CPD methods have been proposed recently, three unresolved challenges remain: (a) poor scalability; (b)  difficulty detecting subtle-change-points;  and (c) intolerance to outliers. Challenge (a) must be tackled since a data stream potentially has an infinite number of data points. A method, which is unable to deal with challenge (b), often misses important subtle change-points in a data stream exactly because they are difficult to detect. A method, which is intolerant to outliers, is unsuitable in the real world as it produces too many false alarms since outliers are common in data streams.

Traditional point-regression-based methods, such as \emph{Autoregressive Moving Average} (ARMA)~\shortcite{box2013box}, 
\emph{Autoregressive Gaussian Process} (ARGP)~\shortcite{candela2003propagation} and 
\emph{ARGP-BOCPD}~\shortcite{saatcci2010gaussian}, usually cannot meet the challenge (c). They detect the change-points by computing the gap between the true values and the predicted values of time series data. Since  outliers are usually far away from the predicted normal value, these methods would inevitably misidentify them as change-points. Other regression-based methods are based on interval pattern matches, such as \emph{FLOSS}~\shortcite{gharghabi2019domain} and \emph{ESPRESSO}~\shortcite{deldari2020espresso}. However, they usually have a high time complexity and do not meet the challenge (a), thus, they are unable to deal with large data streams.

Recent deep learning-based methods \shortcite{cho2014learning,lai2018modeling,deldari2021time} also do not meet the challenge (a) due to training efficiency, although they are powerful enough to handle unstructured and multivariate data streams. They generally require a large amount of data with heavy training costs.  

Moreover, distribution-based methods~\shortcite{candela2003propagation,saatcci2010gaussian,yamada2013change,liu2013change,li2015m} usually have good scalability and tolerate outliers based on kernel methods, which are effective and efficient for solving difficult machine learning tasks. However, existing kernel-based CPD methods usually cannot meet the challenge (b), since they use Gaussian distributional kernel as a means to compute the similarity between distributions of intervals.

Using Gaussian distributional kernel has two key limitations: 
\begin{enumerate} 
    \item The feature map of Gaussian kernel has intractable dimensionality. As a result, measuring the similarity between distributions with Gaussian distributional kernel has a high time cost or it can be sped up via an approximation such as the Nystrom method \shortcite{Nystrom_NIPS2000} to produce a finite-dimensional feature map; 
    \item Gaussian distributional kernel is not sensitive to the subtle change of distributions between two temporal adjacent intervals. 
\end{enumerate}

\begin{figure}[!tb]
\begin{subfigure}{\textwidth}
  \centering
  \includegraphics[width=0.83\linewidth]{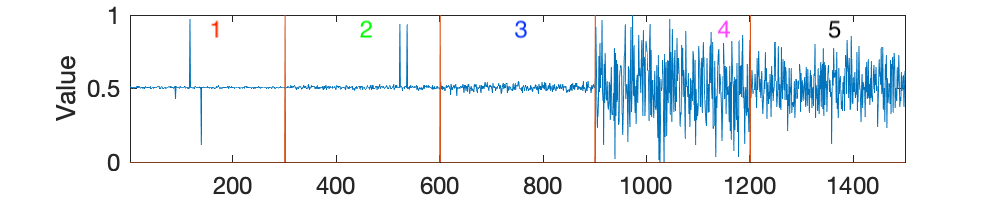}  
  \caption{S1}
  \label{fig:syn-1}
\end{subfigure}

\begin{subfigure}{\textwidth}
  \centering
  \includegraphics[width=0.83\linewidth]{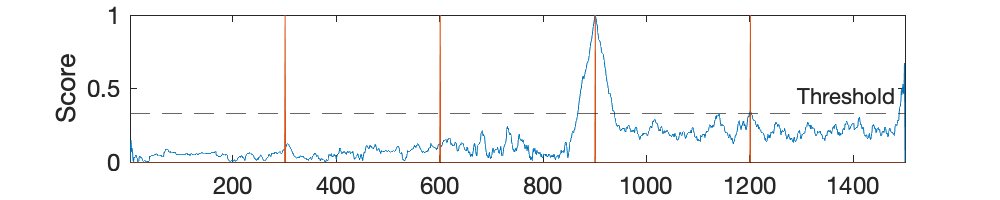}  
  \caption{Using GDK}
  \label{fig:syn1-gdk}
\end{subfigure}

\begin{subfigure}{\textwidth}
  \centering
  \includegraphics[width=0.83\linewidth]{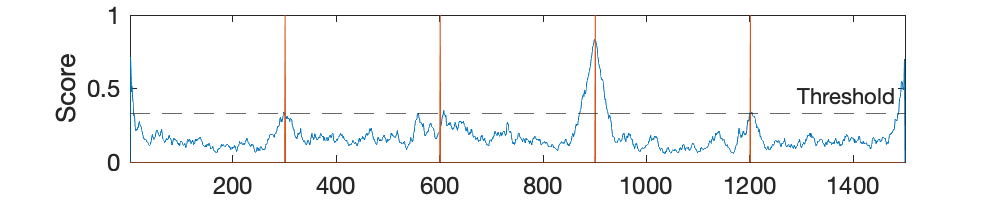}  
  \caption{Using IDK}
  \label{fig:syn1-idk}
\end{subfigure}

\caption{S1 dataset: Comparison of the same proposed distributional kernel-based CID algorithm using GDK vs IDK. (a) shows the data distribution. (b) and (c) plot the change-point score with different kernels. Red bars indicate ground-truth change-points. The variances of five blocks are $1.0, 2.2, 4.3, 48.3$ \& $28.3$, respectively. There are 5 manually added outliers in the first two blocks.}
\label{fig:syn1}
\end{figure}

Figure \ref{fig:syn1} shows the results of CPD on a synthetic dataset using  Gaussian Distributional Kernel (GDK) and Isolation Distributional Kernel (IDK)\footnote{Section \ref{secData} provides the properties of S1 dataset. The details of the proposed CPD methods are provided in Section \ref{sec-idk}. The evaluation results using other kernel methods are given in Appendix~\ref{kcpd}.}. The streaming data contains five different distributions in five blocks (indicated by five numbers on the top) and five outliers in the first two blocks. 

When using Gaussian distributional kernel, shown in Figure \ref{fig:syn1}(b), to measure the dissimilarity between adjacent intervals, the obvious change of distributions between blocks 3 \& 4 shows a high score. However, the change between blocks 4 \& 5 is very subtle; the changes between blocks 1 \& 2 as well as blocks 2 \& 3 cannot be detected.

\begin{figure}[!tbp]
    \centering
    \begin{subfigure}{0.45\textwidth}
        \centering
        \includegraphics[width=0.68\textwidth]{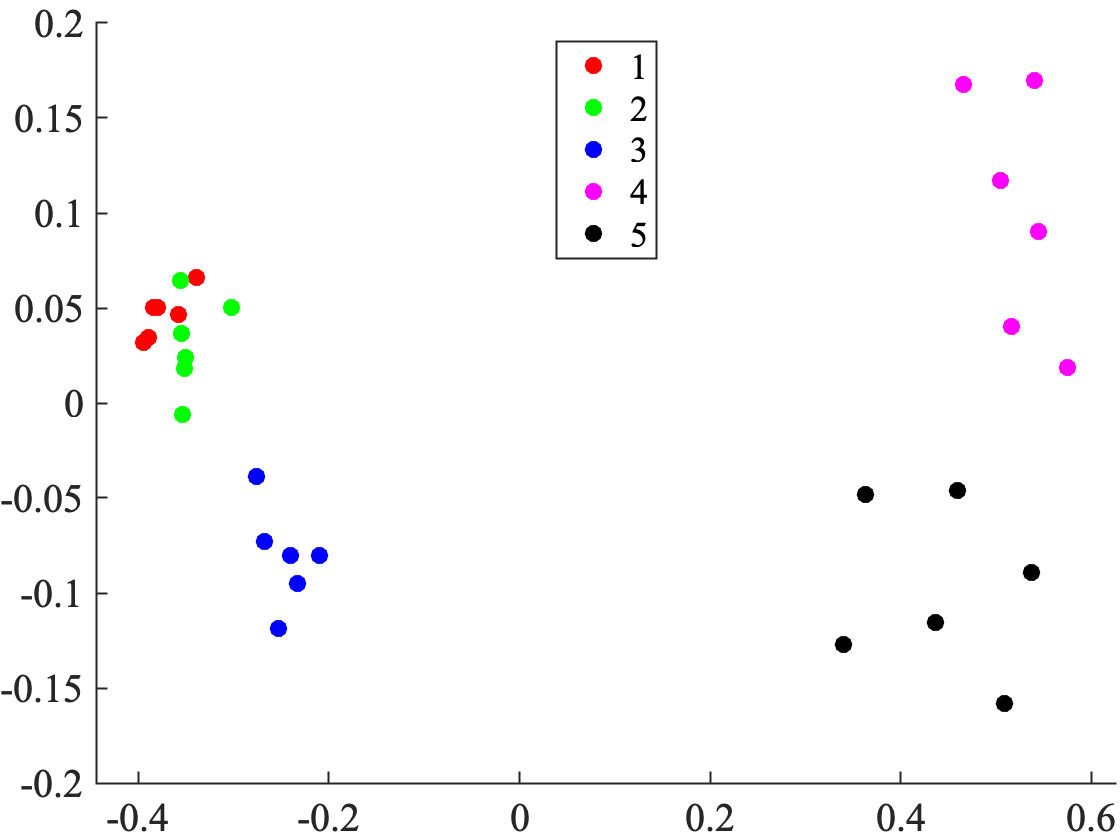}
        \caption{Using GDK}
        \label{fig_compare1b}
    \end{subfigure}
    \begin{subfigure}{0.45\textwidth}
        \centering
        \includegraphics[width=0.68\textwidth]{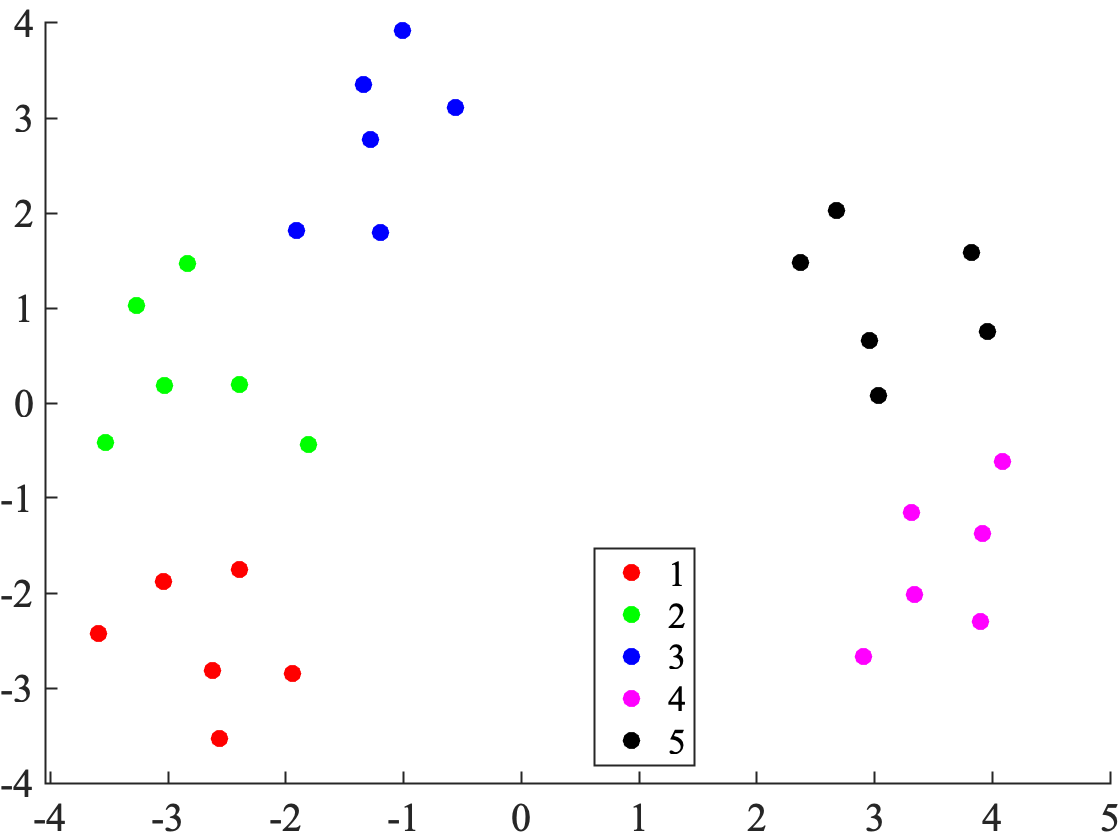}
        \caption{Using IDK}
     \label{fig_compare1a}
    \end{subfigure}    
    \caption{MDS result based on the dissimilarity matrix of intervals of S1 dataset. The split intervals are shown as points.}
    \label{fig_compare1}
\end{figure}

To meet the existing challenges of CPD methods, we propose an \textbf{I}solation Distributional Kernel-based \textbf{C}hange \textbf{I}nterval \textbf{D}etection (\texttt{iCID}) algorithm.
Compared with existing CPD methods, \texttt{iCID} has the following key advantages:
\begin{enumerate}
    \item Able to identify three types of change-points in data streams, which is defined in Section~\ref{sec_change-point}. Figure \ref{fig:syn1-idk} shows that the proposed method can detect type I subtle-changes that are missed by using Gaussian distributional kernel. 
    Figure \ref{fig_compare1} visualises the dissimilarity scores between adjacent intervals based GDK and IDK using MDS.\footnote{Multidimensional scaling (MDS)~\shortcite{borg2012applied} performs a transformation from a high-dimensional space to a $2$-dimensional space for visualisation by preserving as well as possible the pairwise global distances in both spaces. We first split S1 data into intervals and then calculate the distributional dissimilarity between different intervals based on IDK and GDK. The split intervals are shown as points in Figure \ref{fig_compare1}.}
    It shows that intervals from different blocks are more uniformly distributed with IDK-based dissimilarity, and the first two intervals are easier to separate using IDK than using GDK. 
   \item Robust to outliers. The change score is calculated based on distribution/interval rather than a single point. Figure \ref{fig_compare1} shows that all five manually added outliers have low changing scores regardless of using GDK or IDK. 
    \item Linear time complexity of both offline and online versions. The offline version of \texttt{iCID} with a parameter optimisation process can identify change intervals on a streaming dataset with 100,000 points in a few minutes. Furthermore, the online version takes only a few seconds on the same dataset. 
\end{enumerate}

The contributions of this paper are:

\begin{enumerate}
    \item Defining three main types of change-points  in  data streams. 
    \item  Generalising the CPD problem as a special case of Change-Interval Detection (CID) problem. Although all existing works focus on the CPD problem, we find it more efficient and effective to identify change intervals instead.
    \item Proposing IDK-based algorithm \texttt{iCID} for change interval detection in streaming datasets. We are the first to adopt the recent Isolation Distributional Kernel to measure the similarity between adjacent intervals. The source code of \texttt{iCID}  can be obtained from \url{https://github.com/IsolationKernel/iCID}.
    \item Verifying the effectiveness and efficiency of \texttt{iCID} on synthetic and real-world datasets. Our empirical results show that \texttt{iCID} performs better than 6 state-of-the-art methods, including deep learning methods in terms of F1-score and/or runtime. 
\end{enumerate}

The rest of the paper is organised as follows. 
We first describe the related work of CPD in Section~\ref{related-work} and then define the problem for change-interval detection in Section~\ref{prob}. 
Section~\ref{sec-idk} introduces IDK for measuring interval similarity and the proposed \texttt{iCID} algorithm. 
The experimental settings and results are presented in Section~\ref{sec-experiment} and Section~\ref{sec-result}, respectively. 
Finally, we provide a discussion in Section~\ref{sec-discussion}, 
followed by a conclusion in Section~\ref{sec-conclusion}.

\section{Related Work}~\label{related-work}

\begin{table}  
\centering
	\renewcommand{\arraystretch}{1}
	\setlength{\tabcolsep}{2pt}
  \caption{A Comparison of CPD Methods. $n$ in time complexity is the size of the window or the number of observations.} 
  \label{tbl:methods_comparison}
 \small
  \begin{tabular}{llcccc}
\toprule
    Method  & Type  & \specialcell{Time\\Complexity} & \specialcell{Feature\\Learning}  & Core Method & \specialcell{Distribution\\focus}\\
\midrule
    \emph{ARMA}~\shortcite{box2013box}  & Point-regression & $O(n)$ & No  
    & Moving Average & No\\
    \empty{ARGP}~\shortcite{candela2003propagation}  & Point-regression & $O(n^3)$ & No   & Kernel-based & Yes\\
    \emph{ARGP-BOCPD}~\shortcite{saatcci2010gaussian} & Point-regression & Polynomial & No   & Kernel-based & Yes\\
    \emph{FLOSS}~\shortcite{gharghabi2019domain} & Point-regression & $O(nlog(n))$ & No   & Shape-based & No\\
    \emph{ESPRESSO}~\shortcite{deldari2020espresso} & Point-regression & $O(n^2)$ & No   & Shape-based & No\\
    \emph{aHSIC}~\shortcite{yamada2013change} & Distribution & $O(n^2)$ & No  & Kernel-based & Yes\\
    \emph{RuLSIF}~\shortcite{liu2013change}  & Distribution & Unstated   & No  & Kernel-based & Yes\\
    \emph{Mstat}~\shortcite{li2015m}  & Distribution & $O(n)$ & No  & Kernel-based & Yes\\
    \emph{e-divisive}~\shortcite{matteson2014nonparametric} & Distribution & $O(n^2)$ & No  & Statistical-based & Yes
    \\ 
    \emph{KLIEP}~\shortcite{kawahara2009change} & Distribution & $O(n^2)$& No  & Statistical-based & Yes \\
    \emph{RNN En/Decoder}~\shortcite{cho2014learning}  & Deep Learning & Polynomial & YES  & Supervised & No\\
    \emph{LSTNet}~\shortcite{lai2018modeling}  & Deep Learning & Polynomial & YES  & Supervised & No\\
    \emph{KL-CPD}~\shortcite{chang2019kernel}  & Deep Learning & Polynomial & YES  & Kernel-based & Yes\\
    \emph{TS-CP$^2$}~\shortcite{deldari2021time}  &Deep Learning & Polynomial & YES  & self-supervised & No\\
\hdashline
    \textbf{\texttt{iCID}} & \textbf{Distribution} & \textbf{O(n)} & \textbf{No} & \textbf{Kernel-based} & \textbf{Yes}\\
\bottomrule
\end{tabular}
\end{table}

We summarise the methods for change-point detection in Table~\ref{tbl:methods_comparison} and classify them into the following three subsections.

\subsection{Point-Regression Based CPD}

Point-regression based methods detect the change-points by computing the gap between the true values and the predicted values of time series data. 
Parametric and non-parametric model can be used for time series data regression. 
\emph{Autoregressive Moving Average} (ARMA) \shortcite{box2013box} is a parametric model 
that combines \emph{Autoregressive} (AR) and \emph{Moving Average} (MA) for forecasting time series,
where the AR describes the relationship between current and historical values. 
The MA generates a linear model of the error accumulation of the autoregressive part.
Parametric models can greatly simplify the process of model learning 
because they specify the form of the objective function. 
However, 
they usually do not exactly match the underlying objective function, 
which may lead to underfitting and underperformance.

The non-parametric methods can fit different functional forms without assumptions about the probability distribution.
\emph{Autoregressive Gaussian Process} (ARGP)~\shortcite{candela2003propagation} as a non-parametric model 
can be considered as a nonlinear version of AR based on Gaussian process for time series prediction.
\emph{ARGP-BOCPD}~\shortcite{saatcci2010gaussian} uses ARGP in underlying predictive models of \emph{Bayesian online change-point detection} (BOCPD) framework to extend BOCPD. 
There are other methods extracting the shape patterns of time series.
\emph{FLOSS}~\shortcite{gharghabi2019domain} identifies the position of change-points by comparing changes in interval shape pattern.
\emph{ESPRESSO}~\shortcite{deldari2020espresso} combines shape patterns (based on FLOSS) and the statistic approach (information-gain based~\shortcite{sadri2017information}) to identify change-points to detect potential segment boundaries.  

However, since outliers are usually far away from the predicted value,
point-regression based methods would inevitably misidentify them as change-points when they set a threshold on the gap between the test value and the predicted value. Moreover, the shape-based~\shortcite{gharghabi2019domain,deldari2020espresso} approaches usually have a high time complexity for intervals matching.

\subsection{Distribution-Based CPD}

Distribution-based CPD methods detect change-points by identifying the difference between two distributions via some statistic, where points in each consecutive interval in a data stream are assumed to be a set of i.i.d points generated from some unknown distribution. They have shown promising performance in CPD tasks~\shortcite{yamada2013change,liu2013change,li2015m}. These methods are usually non-parametric based kernel methods, thus, they are more robust to outliers in providing consistent results on different types of data distributions than point-regression based CPD algorithms.

\emph{aHSIC}~\shortcite{yamada2013change} uses Gaussian distributional kernel to calculate the dissimilarity of past and future interval data separability to detect change-points.
The method \emph{Mstat} proposed in~\shortcite{li2015m} uses \emph{M-statistics} to detect change-points based on Gaussian distributional kernel \emph{Maximum Mean Discrepancy} (MMD) for a two-sample test between the current interval distribution with a reference interval distribution having no change-points.
\emph{RuLSIF}~\shortcite{liu2013change} uses a density-ratio estimation for CPD 
and uses Gaussian distributional kernel to model the density-ratio distributions between intervals of time series.
However, existing GDK-based CPD methods have limited sensitivity to subtle changes, as they rely on a data-independent kernel that cannot adapt to the varied density of the data. 
Compared to GDK-based CPD methods, IDK has a finite-dimensional feature map, and the data-dependent property of IDK makes \texttt{iCID} easier to detect subtle change-points. 

Another way to measure the dissimilarity between distributions is to use a statistical-based method. \emph{e-divisive}~\shortcite{matteson2014nonparametric} uses a U-statistic test to measure the dissimilarity between the distributions of two adjacent intervals to identify change-points, but it has a time complexity $O(n^2)$ that cannot handle the large scale of data. \emph{KLIEP}~\shortcite{kawahara2009change} detects change-points by using the Kullback Leibler (KL) divergence to estimate the probability of density between two intervals. However, \emph{KL-divergence} is sensitive to outliers or noise.

In summary, a general approach for distribution-based CPD methods is to treat each interval as a distribution and use a measure to compute the similarity between them. The main difference among these methods lies in the choice of the similarity measure.

\subsection{Deep Learning-Based CPD}

Deep learning based CPD methods have been developed in recent years.
\emph{RNN Encoder-Decoder}~\shortcite{cho2014learning} uses two \emph{Recurrent Neural Networks} (RNN) 
as encoder and decoder respectively 
to learn the nonlinear features of the interval
and forecast time series.
\emph{LSTNet}~\shortcite{lai2018modeling} is a deep learning framework designed 
for multivariate time series forecasting tasks, 
which combines \emph{RNN} and \emph{CNN} to forecast future time series 
by extracting short-term local dependency patterns between variables.
\emph{KL-CPD}~\shortcite{chang2019kernel} uses kernel methods and two-sample tests 
to measure differences between continuous intervals 
to detect change-points. Its definition assumes i.i.d., yet the time dependency is incorporated in the generative modelling using \emph{RNN} to model the changed distribution which is close to, but different from, the distribution before the change. The modelled distribution is a surrogate distribution in order to generate points to simulate points from a changed distribution where points are few (or unavailable) in practice.
\emph{TS-CP$^2$}~\shortcite{deldari2021time} uses a single positive pair of 
contiguous time windows and a set of negative separated across time windows
to train the encoder with contrastive learning. 
\emph{TS-CP$^2$} also used \emph{Wavenet} to 
learn representation between time intervals 
and hypothesised that a change-point will be presented 
if there is a significant representation difference 
between time-adjacent intervals.

However, deep-learning-based methods are inefficient in handling massive data streams because they usually require a large amount of data for expensive training. 
Thus, they cannot be used in the online setting.

 \hspace{1cm}

\section{Problem Definition} 
\label{prob}

\subsection{Change-Point Detection}
\label{sec_change-point}
Existing works \shortcite{chang2019kernel} define a change-point as a time step when the data distribution before the change-point is different from the data distribution after the change-point, i.e., the data distribution has an abrupt change after the change-point.

Let $\mathcal{P}^t_L$ and $\mathcal{P}^t_R$ be the distributions before and after a point $x_t$,  respectively, in a series of observations $\{x_1, x_2, x_3,...\}$, where each distribution is estimated from $w$ observations before and after $x_t$ which are assumed to be their  i.i.d. samples.

\begin{definition}\label{changePoint}
Given a series of observations $\{x_1, x_2, x_3,...\}$, 
$x_t$ is a 
\begin{itemize}
    \item \textbf{change-point}  if $\mathcal{P}^t_L \ne \mathcal{P}^t_R$, and either $x_t \sim \mathcal{P}^t_L$ or  $x_t \sim \mathcal{P}^t_R$;
    \item \textbf{same-distribution point}  if $\mathcal{P}^t_L = \mathcal{P}^t_R$, and $x_t \sim \mathcal{P}^t_L$ or  $x_t \sim \mathcal{P}^t_R$;
    \item \textbf{outlier} if $\mathcal{P}^t_L = \mathcal{P}^t_R$, and neither $x_t \sim \mathcal{P}^t_L$ nor $x_t \sim \mathcal{P}^t_R$.
\end{itemize}
\end{definition}

The small difference in distributions can be further defined as \textbf{subtle-change}: $x_t$ is a subtle-change-point if the difference between $\mathcal{P}^t_L$ and $\mathcal{P}^t_R$ is small.

There are mainly three types of change-points defined as follows:

\begin{enumerate}[label=(\Roman*)]
    \item Sudden change: the distribution is changed suddenly after the change-point, i.e., there is only one change-point in the interval. Many existing CPD methods \shortcite{yamada2013change,li2015m,chang2019kernel} focus on this type of change by identifying the change-point defined in Definition \ref{changePoint}.
    \item Gradual change: a distribution is gradually changed to another distribution over a short period in which points generated from two different distributions co-exist, i.e., $x$ is generated from a mixture of $\mathcal{P}_L$ and $\mathcal{P}_R$ within the period. The distributions before and after the period are $\mathcal{P}_L$ and $\mathcal{P}_R$, respectively.  
    \item Continuous change: a distribution keeps changing, potentially, over a period, i.e., many points $x_t$ within the period, $\mathcal{P}^t_L \ne \mathcal{P}^t_R$. 
\end{enumerate}

In real-world applications, 
identifying every single change-point in a data stream is impractical and unnecessary due to two main reasons: 
\begin{enumerate}
    \item It is time-consuming for any method to check every point in a data stream, especially since the data stream can be infinite; 
    \item  In the period of gradual change or continuous change, there are many change-points. 
\end{enumerate}

\subsection{Change-Interval Detection} \label{cid}

Here we argue that the CPD problem can be better treated when it is defined as a \emph{change-interval detection} (CID) problem, where the detection focus is the interval that contains at least one change-point. In addition, the CPD problem is a special case of CID problem. 
Most of the existing works on CPD focus on identifying the exact locations of changes in a dataset. However, we argue that it is more efficient and effective to detect change intervals instead, as many datasets have inaccurate labels that can affect the performance of point-based methods significantly. In contrast, interval-based methods can handle such label issues more robustly because it detects the intervals that cover the change-points instead of the exact location of the change-points. 

\begin{definition}  
A \textbf{change interval} contains one or more change-points.
\label{def-change-interval}
\end{definition} 

Thus, the distribution in the change interval is different from that in the previous interval. Given a distributional measure $f(\cdot)$ to calculate the similarity between two distributions, a $f_{\tau}$-change interval is formally defined as follows.  

\begin{definition} Given two temporal adjacent intervals $L$ and $R$, each having length $w$,
 $R$ is a $f_{\tau}$-change interval if the similarity between the two distributions $\mathcal{P}_L$ and $\mathcal{P}_R$  is less than a threshold $\tau$:
\begin{eqnarray}
    f(\mathcal{P}_L,\mathcal{P}_R) \le \tau.
\end{eqnarray}   
\end{definition}

Different distributional measures have been developed to 
compute the similarity between two intervals, e.g., 
 \emph{KL (Kullback–Leibler) divergence}~\shortcite{csiszar1975divergence},
\emph{JS (Jensen-Shannon) divergence}~\shortcite{manning1999foundations},
\emph{Wasserstein metric}~\shortcite{vaserstein1969markov},
and \emph{MMD}~\shortcite{MMD-2006}.

We propose a generic method for CID which can be converted to detect CPD by using sliding windows.
The details of the proposed CID and CPD method are provided in Section \ref{sec-idk} and \ref{sectionCPD}, respectively. Nevertheless, both methods overcome the three unresolved challenges (a), (b) \& (c) stated in Section \ref{intro}.

\section{Change-interval Detection Based on Distributional Kernel}~\label{sec-idk}

We first provide the pertinent details of distributional kernels, used for measuring similarities between distributions, in the first subsection. Then, we propose a new algorithm for change-interval detection called \texttt{iCID}, based on the distributional kernel, in the second subsection. The third subsection introduces the automatic parameter selection for offline and online versions of \texttt{iCID} for real practice.

\subsection{Distributional Kernels}
\label{sec-IDK}

Given two probability distributions $\mathcal{P}_X$ and $\mathcal{P}_Y$, the similarity between them can be measured using a distributional kernel.
\begin{definition}
A \textbf{distributional kernel}, based on Kernel Mean Embedding (KME)~\shortcite{muandet2017kernel}, is given as:
\begin{eqnarray} 
    \mathcal{K}(\mathcal{P}_X, \mathcal{P}_Y) & = & 
    \frac{1}{|X||Y|}\sum_{x\in X} \sum_{y\in Y} \kappa (x, y) \nonumber \\ 
     & = & \left<\widehat{\Phi}(\mathcal{P}_X),\widehat{\Phi}(\mathcal{P}_Y) \right>,
    \end{eqnarray}
\end{definition}

\noindent where $\widehat{\Phi}(\mathcal{P}_X)  =  \frac{1}{|X|} \sum_{x \in X} \Phi(x)$ is the feature map of the kernel mean embedding.

KME often uses Gaussian kernel as $\kappa$; and we call it Gaussian Distributional kernel (GDK) in this paper.

Recently, a new \emph{Isolation Distributional Kernel} (IDK) has been  proposed to 
perform point \& group anomaly detection~\shortcite{ting2022isolation} 
and time series anomaly detection~\shortcite{ting2022new}.

The key idea of the Isolation kernel is using a space partitioning strategy to split the whole data space into $\psi$ non-overlapping partitions based on a random sample of $\psi$ points from a given dataset. The similarity between any two points is how likely these two points can be split into the same partition.

Let $D \subset \mathcal{X} \subseteq \mathbb{R}^d$ be a dataset sampled from an unknown probability distribution $\mathcal{P}_D$; 
Additionally,
let $\mathds{H}_\psi(D)$ denotes the set of all partitionings $H$ admissible from the given dataset $\mathcal{D} \subset D$,
where each point $z \in \mathcal{D}$ has the equal probability of being selected from $D$.
Each of the $\psi$ isolating partitions $\theta[{ \mathbf{z}}] \in H$ isolates one data point ${ \mathbf{z}}$ 
from the rest of the points in a random subset $\mathcal D$, 
and $|\mathcal D|=\psi$. 
Each $H$ denotes a partitioning from each $\mathcal D$~\shortcite{qin2019nearest}. 

\begin{definition}\label{IKernel} 
For any two points $x,y \in \mathbb{R}$,
	Isolation Kernel of $x$ and $y$ is defined to be
	the expectation taken over the probability distribution on all partitionings $H \in \mathds{H}_\psi(D)$ that both $x$ and $y$  
    fall into the same isolating partition $\theta[z] \in H$, 
    $z \in \mathcal{D} \subset D$, 
    where $\mathds{1}(\cdot)$ be an indicator function:
	\begin{eqnarray}
        \kappa_I(x,y\ |\ D)  & = & {\mathbb E}_{\mathds{H}_\psi(D)} [\mathds{1}(x,y \in \theta\ | \ \theta \in H)]. \nonumber\\
         \label{Eqn_kernel}
	\end{eqnarray}
In practise, 
$\kappa_I$ is constructed using a finite number of partitionings $H_i, i=1,\dots,t$, (t=200 as default)
where each $H_i$ is generated by using randomly subsampled $\mathcal{D}_i \subset D$; 
$\theta$ is a shorthand for $\theta[z]$; 
$\psi$ is the sharpness parameter:
    \begin{eqnarray}
        \kappa_I(x,y\ |\ D)  & = & \frac{1}{t} \sum_{i=1}^t   \mathds{1}(x,y \in \theta\ | \ \theta \in H_i).
        \label{Eqn_IK}
    \end{eqnarray}
\end{definition}
\noindent

\begin{figure}[!hb]
  \centering
\begin{subfigure}{0.48 \textwidth}
  \centering
  \includegraphics[width=1\linewidth]{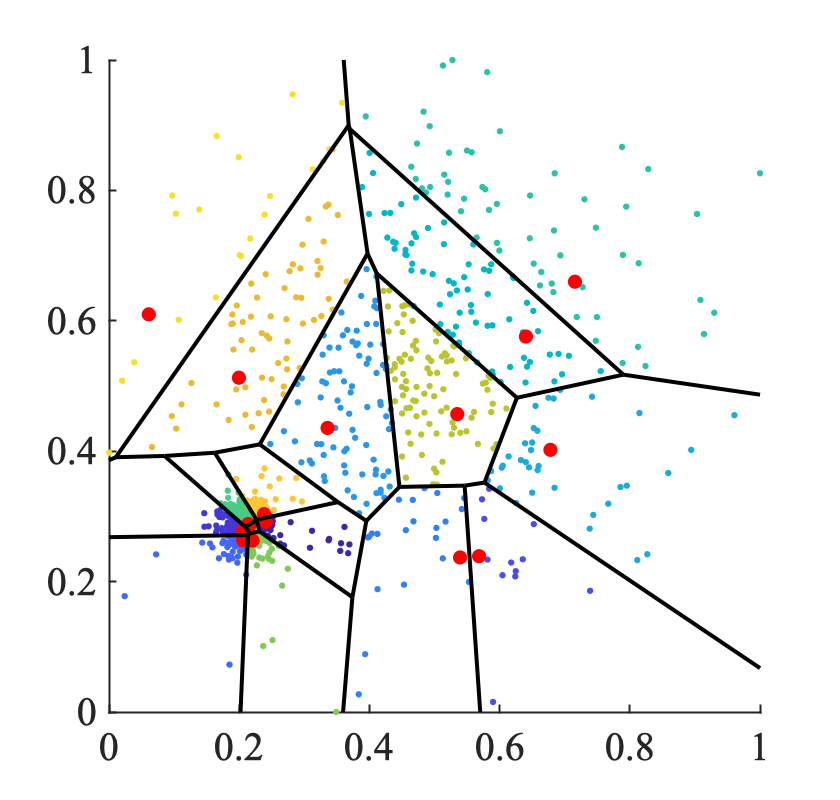}  
  \caption{Sample Partitioning}
  \label{fig:vonoroi}
\end{subfigure}
\begin{subfigure}{0.48 \textwidth}
  \centering
  \includegraphics[width=1\linewidth]{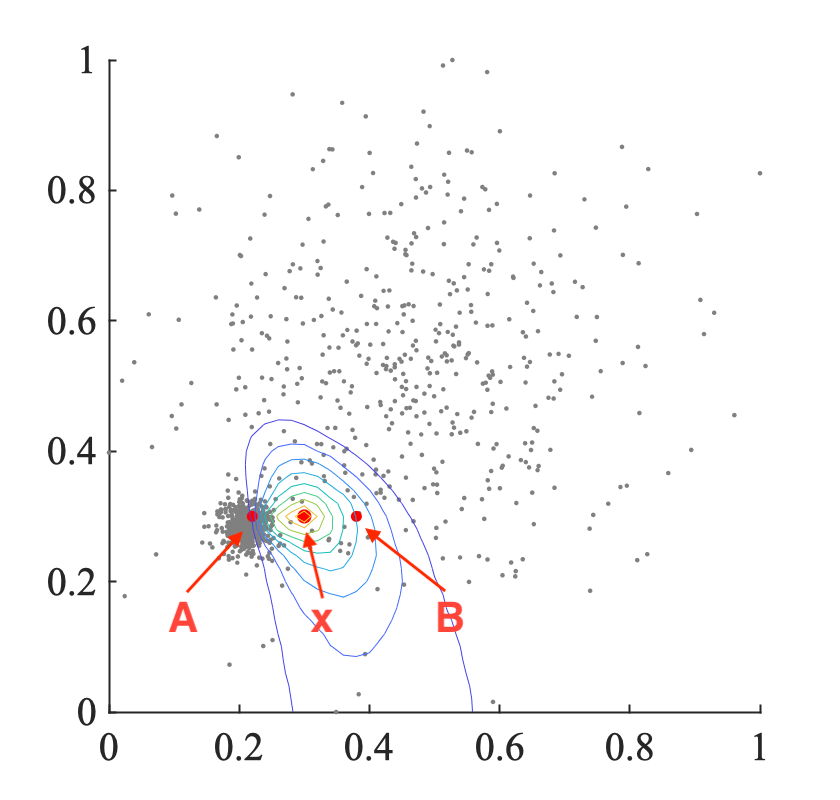}  
  \caption{Contour}
  \label{fig:cont}
\end{subfigure}
\caption{Illustration of the similarity calculation using Isolation kernel ($\psi$ = 16): (a) An example partitioning $H$. (b) Contours with reference to point $x = (0.3,0.3)$.}
\label{fig:partitioning}
\end{figure}

In this paper, we use the Voronoi diagram~\shortcite{aurenhammer1991voronoi} to partition the space, i.e., each $H \in \mathds{H}_\psi(D)$ is a
Voronoi diagram and each sample point $z \in \mathcal{D}$ is the centre of each cell in Voronoi diagram. We illustrate a partitioning of $H$ on the S2 dataset with $\psi = 16$ in Figure~\ref{fig:vonoroi}, i.e, the data space is partitioned into 16 cells based on the 16 red points.

Figure~\ref{fig:vonoroi} shows that the dense region is partitioned into smaller cells than the sparse region. Consequently, data points in the dense region (being in smaller cells) have lower similarity scores than data points with the same inter-point distance  in the sparse region (being in the larger cells).
Figure~\ref{fig:cont} illustrates this phenomenon, such that data point $x = (0.3,0.3)$ is more similar to $B = (0.38, 0.3)$ from a sparse region than $A = (0.22, 0.3)$ from a dense region, although $x$ has the same distance to both $A$ and $B$.

Isolation kernel has a finite feature map, which is defined as follows~\shortcite{ting2020clustering,ting2020isolation}:
\begin{definition}
	\label{def:featureMap}  \textbf{Feature map of Isolation Kernel.}
	For point $x \in \mathbb{R}$, the feature mapping $\Phi: x\rightarrow \mathbb \{0,1\}^{t\times \psi}$ of $\kappa_I$ is a vector that represents the partitions in all the partitioning $H_i\in \mathds{H}_\psi(D)$, $i=1,\dots,t$; where $x$ falls into either only one of the $\psi$ hyperspheres in each partitioning $H_i$.
\end{definition}

Re-express Equation (\ref{Eqn_IK}) using $\Phi$ is shown as follows:
\begin{eqnarray}
    \kappa_I(x,y\ |\ D) = \frac{1}{t} \left< \Phi(x|D), \Phi(y|D) \right>.
    \label{eqn_IK_feature_map}
\end{eqnarray}

Given the feature map $\Phi$ (defined in Definition~\ref{def:featureMap}) and Equation~(\ref{eqn_IK_feature_map}), 
the estimation of KME can be expressed based on the feature map of Isolation Kernel $\kappa_I(x,y)$.

Then IDK is defined as 
follows \shortcite{ting2020isolation}:
\begin{definition}\label{IDKernel}
Isolation Distributional Kernel of two distributions $\mathcal{P}_X$ and $\mathcal{P}_Y$ is shown as follows: 
\begin{eqnarray}
{\mathcal{K}}_I(\mathcal{P}_X,\mathcal{P}_Y\ |\ D) 
 & = & \frac{1}{t} \left< \widehat{\Phi}(\mathcal{P}_X|D), \widehat{\Phi}(\mathcal{P}_Y|D) \right> ,\label{eqn_IDK}
\end{eqnarray}
where  $\widehat{\Phi}(\mathcal{P}_X|D)  =  \frac{1}{|X|} \sum_{x \in X} \Phi(x|D)$ is the empirical feature map of kernel mean embedding. 
 \end{definition}

Note that we normalise the IDK similarity to $[0, 1]$ as follows:

\begin{equation} 
	\widehat{\mathcal{K}}_I(\mathcal{P}_X,\mathcal{P}_Y\ |\ D)
	 =  \frac{\langle  \widehat{\Phi}(\mathcal{P}_X|D), \widehat{\Phi}(\mathcal{P}_Y|D) \rangle}{\sqrt{\langle  \widehat{\Phi}(\mathcal{P}_X|D), \widehat{\Phi}(\mathcal{P}_X|D) \rangle}\sqrt{\langle \widehat{\Phi}(\mathcal{P}_Y|D), \widehat{\Phi}(\mathcal{P}_Y|D) \rangle}}.
	\label{E5}
\end{equation}

IDK has two distinctive characteristics in comparison with GDK. First,
IDK has a finite-dimensional feature map 
that enables efficient computation of distributional similarity with linear time complexity. In contrast, GDK has a feature map with infinite-dimensionality. Second, IDK is a data-dependent measure, i,e., two distributions, as measured by IDK derived in a sparse region, are more similar than the same two distributions, as measured by IDK derived in a dense region. GDK is a data-independent  kernel.

An example of the data-dependent property is  shown in Figure \ref{fig:syn1}: the dissimilarity between blocks 1 and 2 is enlarged; but the dissimilarity between blocks 4 and 5 is reduced. Therefore, the dissimilarity scores between different adjacent blocks reach to similar level such that subtle change-points in the original data stream become globally obvious change-points that are easier to be detected using a single threshold, as shown in Figure \ref{fig:syn1-idk}.

In this paper, 
we focus on using IDK as the distributional measure $f$
to compute the similarity between two temporal adjacent intervals. To be consistent with other existing measures of difference or discrepancy, 
we use $\mathfrak{S}(X, Y)=1- f(\mathcal{P}_X,\mathcal{P}_Y)$
as the dissimilarity score between two intervals $X$ and $Y$ hereafter.

\begin{figure}[!tb]
    \centering
    \includegraphics[width = 0.45\textwidth]{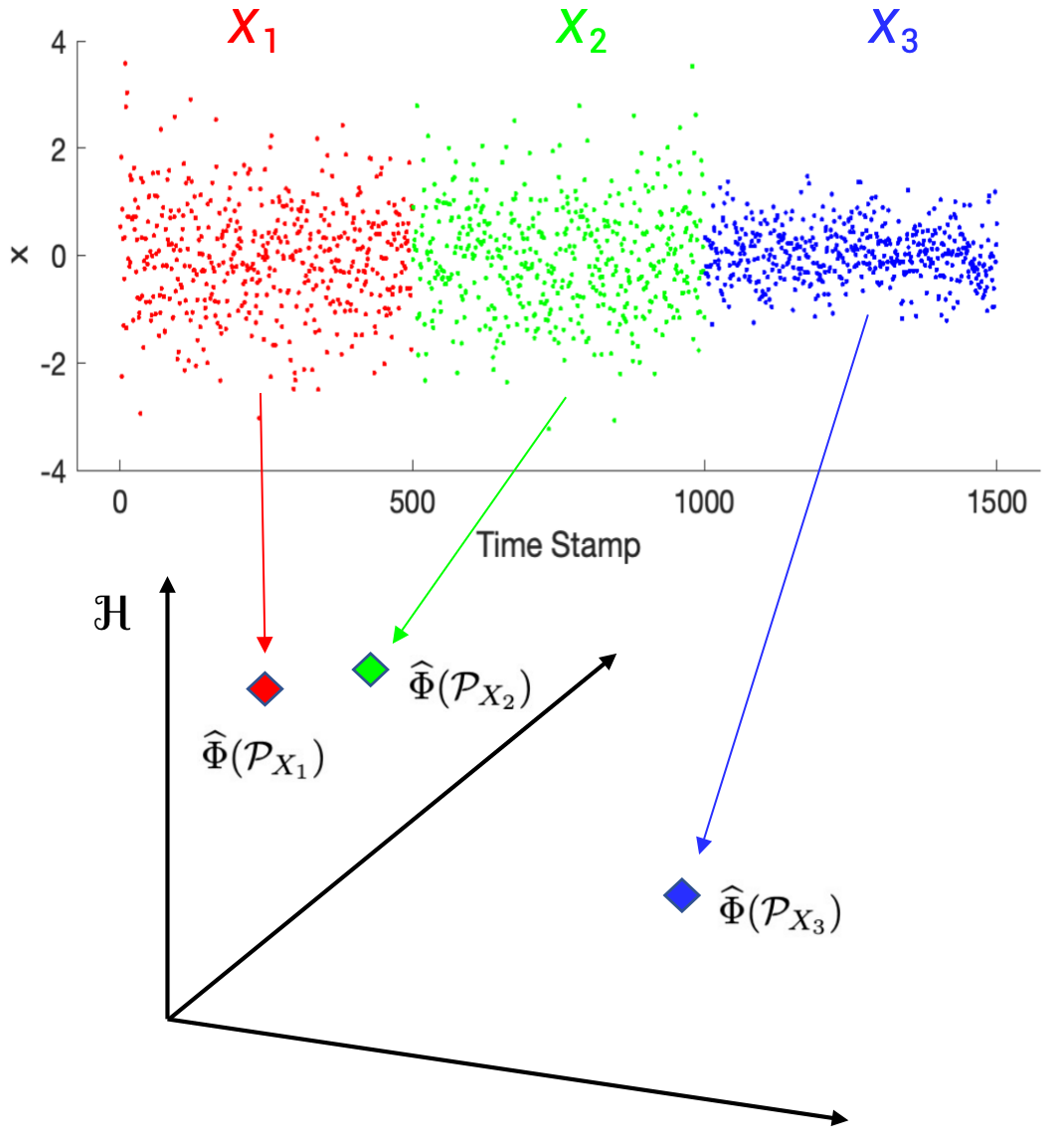}
    \caption{Feature mapping of each time interval $X_i$ as a distribution to a point $\widehat{\Phi}(\mathcal{P}_{X_i})$ in the feature space of IDK or KME. Similar distributions are mapped into the same region of the feature space; and different distributions are mapped into different regions. }
    \label{fig:kme}
\end{figure}

\subsection{The Proposed $\texttt{iCID}$}
 
The main idea of \textbf{I}solation
Distributional Kernel-based \textbf{C}hange \textbf{I}nterval \textbf{D}etection (\texttt{iCID}) is to treat each interval in a data stream as a distribution and use IDK as the measure to calculate the dissimilarity between distributions of an interval and its earlier adjacent interval. Each interval is assigned an iCID score based on this dissimilarity, if the score exceeds a threshold, the interval is detected as a change interval.

\begin{figure}[!tb]
  \centering
  \includegraphics[width=0.6\linewidth]{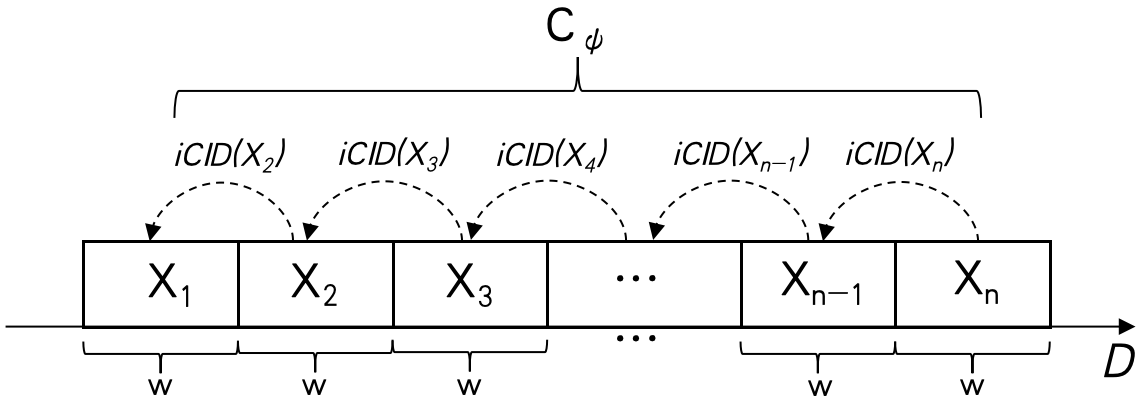}  
  \label{fig:offlinestep}
\caption{Illustration of \texttt{iCID} calculation.
}
\label{fig:step}
\end{figure}

The key procedures of \texttt{iCID} are as follows.
It first splits the dataset $D$ into $N$ non-overlapping time intervals, i.e., $X_1, X_2,...X_N$, where each $X_i$ has the same window size $w$. 
Then it uses IDK to map each time interval $X_i$ into a point $\widehat{\Phi}(\mathcal{P}_{X_i})$ in the feature space of IDK which is shown in Figure~\ref{fig:kme}, and computes the dissimilarity between intervals $X_i$ and its adjacent interval $X_{i-1}$, as illustrated in Figure \ref{fig:step}. Let $C_{\psi}=\{\texttt{iCID}(X_1), $ $\texttt{iCID}(X_2), ..., \texttt{iCID}(X_N) \ | \ \psi, D \}$ be the set of all scores of $N$ intervals in $D$, obtained using $\psi$, a parameter for $\texttt{iCID}$. 
The \texttt{iCID} score of interval $X_i$ is: 
\begin{equation}
\label{IDKCPD}
\texttt{iCID}(X_i)=\mathfrak{S}(X_i, X_{i-1})=1-\widehat{\mathcal{K}}_I(\mathcal{P}_{X_i},\mathcal{P}_{X_{i-1}} \ | \ D ).    
\end{equation}
where $\widehat{\mathcal{K}}_I(\mathcal{P}_{X_i},\mathcal{P}_{X_{i-1}}\ | \ D)$ is the normalised IDK similarity between the distributions of two adjacent intervals.

Here we propose two versions of  \texttt{iCID}, i.e., the offline version and the online version. Recall that IDK uses subsamples $\mathcal{D}$ (each having $\psi$ points, as described in Section \ref{sec-IDK}) to build the partitionings $H \in \mathds{H}_\psi(D)$. In the offline version, the subsamples $\mathcal{D}$ are uniformly sampled from the whole dataset $D$ to calculate the \texttt{iCID} scores for all intervals. However, for the online version with limited memory, $\mathcal{D}$ is sampled from the latest $k$ data points in order to compute the \texttt{iCID} score for the current interval. This is because the latest data points are the best representation of the current interval.

\subsection{Automatic Parameter Selection}

Since IDK has a key parameter $\psi$, here we propose a best $\psi$ selection method as follows. 
Assuming that change-points are rare, the \texttt{iCID} scores in a data stream would have a few large scores only while others have small scores, i.e., the \texttt{iCID} scores are stable overall. 
Therefore, we aim to search for the best parameter $\psi$ in IDK in order to produce the most stable \texttt{iCID} scores.

Let $C_{\psi}=\{\texttt{iCID}(X_1), $ $\texttt{iCID}(X_2), ..., \texttt{iCID}(X_N) \ | \ \psi, D \}$, the best $\psi$ leads to a maximum of stability or a minimum of instability, i.e.,

\begin{equation}
    \psi^* = \mathop{\mathrm{argmin}}\limits_{\psi}{\Bar{E}(C_{\psi})},
        \label{eqn:psi}
\end{equation}

\noindent where $\Bar{E}(.)$ is a measure of instability, e.g., approximated entropy estimation~\shortcite{pincus1991approximate}, variance or Gini coefficient~\shortcite{gini1912variabilita}.\footnote{The following experiment results are obtained by using the  approximate entropy ~\shortcite{pincus1991approximate}. We find that the results of using the other two measurements are similar. The details of the comparison are given in Appendix~\ref{measurement}.} Note that a lower $\Bar{E}(.)$ value means a higher stability.

Once obtaining the iCID score for each interval, we can set a threshold $\tau$ to identify the changing interval $X_j$ with $\texttt{iCID}(X_j)>\tau$. A threshold $\tau$ can be identified as 
\begin{equation}
    \tau(C_{\psi^*}) = \mu + \alpha \times \sigma, 
    \label{eqn:threshold}
\end{equation}
where $\mu$ and $\sigma$ are the mean and standard deviation of $C_{\psi^*}$, respectively; and $\alpha$ is a power factor parameter.

 For the offline version, $C_{\psi}$ is calculated based on all intervals, and we can search the $\psi^*$ in a reasonable pre-defined range.  
Algorithm \ref{shCHalgorithm} illustrates the steps of offline \texttt{iCID}.

\begin{algorithm}
\caption{\texttt{Offline$\_$iCID(D,w,$\Psi$)}}
\label{shCHalgorithm}
\begin{algorithmic}[1]
\Statex \textbf{Input:} Dataset $D$; 
Window Size $w$; Subsample Size List $\Psi$ 
\Statex \textbf{Output:} $C_{\psi^*}$ - a set of $N$ Interval Scores
\State Split $D$ into $N$ non-overlapping time intervals, each having $w$ points, i.e.,
\Statex $D \rightarrow \{X_i, i=1, \dots ,N\}$, where $N=\lfloor	length(D)/w \rfloor	$
\State Search the best $\psi^*$ from the $\Psi$, i.e.,
\Statex $\psi^* = \mathop{\mathrm{argmin}}\limits_{\psi}{\Bar{E}(C_{\psi})}$
\State \textbf{Return} $C_{\psi^*}$
\end{algorithmic}
\end{algorithm}

For the online version, the $\psi^*$ is obtained based on  a reference dataset $D$, e.g., first $k$ points of the streaming data. For calculating the iCID of a current interval $X_i$, the $\psi^*$ subsamples used to build IDK is obtained from the latest streaming dataset $D'$ with $k$ points ending at $X_i$, such that the feature map keeps updating to adapt the distribution of the current interval. Algorithm \ref{online-set} illustrates the steps of online iCID.

\begin{algorithm}[!htb]
\caption{\texttt{Online$\_$iCID(D,D',w,$\Psi$)}}
\label{online-set}
\begin{algorithmic}[1]
\Statex \textbf{Input:} Reference dataset $D$; Current streaming data $D'$;  
Window Size $w$; Subsample Size List $\Psi$ 
\Statex \textbf{Output:} S - the iCID score of the last interval in $D'$
\State$\psi^*$ = Offline\_iCID($D,w,\Psi$)
\State Let $X_N$ and $X_{N-1}$ be the last two adjacent intervals in $D'$ where each interval has $w$ points 
\State  $S =1-\widehat{\mathcal{K}}_I(\mathcal{P}_{X_N},\mathcal{P}_{X_{N-1}} \ | \ \psi^*, \ D' )$  
\State \textbf{Return} $S$
\end{algorithmic}
\end{algorithm}

Note that the automatic parameter selection does not involve optimisation or learning. In other words, the entire process of \texttt{iCID} requires no optimisation or learning for both the online and offline versions. This is the advantage of most existing kernel and statistical-based change point detection methods.

\section{Experimental Settings}~\label{sec-experiment}

In this section, we introduce the datasets and parameter settings used in our experiments. 

\subsection{Benchmark Datasets} 
In order to demonstrate the effectiveness of the proposed \texttt{iCID} method in a variety of applications, we include 6 real-world and 2 synthetic datasets. Table \ref{tab:datasets} presents the properties of each dataset. 
All datasets are normalised such that every dimension is in the same range of [0, 1] before experiments are conducted.

\subsubsection{Real-world Datasets}

\begin{itemize}
    \item \textbf{Yahoo-16}\footnote{https://webscope.sandbox.yahoo.com/catalog.php?datatype=s} records hardware resource usage during the operation of the PNUTS/Sherpa database (e.g. CPU utilization, memory utilization, disk utilization, network traffic, etc). We directly used the dataset from previous work~\shortcite{chang2019kernel} which chose the $16^{th}$ of a total of 68 representative time series after removing some intervals with duplicate patterns in anomalies. 
    Yahoo-16 dataset had 5 type I change-points. 
    \item \textbf{USC-HAD}\footnote{http://sipi.usc.edu/had}~\shortcite{zhang2012usc} includes 14 subjects and 12 daily activities. Each subject was fitted with a 3-axial accelerometer and 3-axial gyroscope, which are fixed in front of the right hip joint and sampled at 100 Hz. We reused the same dataset from~\shortcite{deldari2021time} which randomly chose 30 activities from the first six participants and stitched the selected recordings together in a random manner. 
    USC-HAD dataset has type I sudden-change-points mentioned in Section \ref{sec_change-point}. 
    \item \textbf{Google-Trend}\footnote{https://github.com/zhaokg/Rbeast/blob/master/Matlab/testdata/googletrend.mat} is a monthly time series data of the Google Search popularity of the word `beach' over the US. This time series is regularly-spaced and the frequency period is 1 year. 
    Google-trend dataset has 2 type I sudden-change-points. 
    \item \textbf{Well\_Log}\footnote{https://github.com/alan-turing-institute/TCPD/tree/master/datasets/well\_log}~\shortcite{oruanaidh1996numerical} contains measurements of the nuclear magnetic response and conveys information about the structure of the rock. The series has been sampled every 6 iterations to reduce the length~\shortcite{van2020evaluation}. 
    Well\_log dataset has 10 type I change-points. In addition, it contains some outliers which are defined in Section~\ref{sec_change-point}. 
    \item \textbf{Weather}\footnote{https://datahub.io/core/global-temp\#readme} is global monthly mean temperature time series data in degrees Celsius from 1880 to 2022. Note that this data does not have labels for change-points, but we can easily observe the change trends from the value distribution along the timeline, i.e., the period of 1,400-2,000 and 2,400-3,288 has type II and III change-points, respectively.
    \item \textbf{HASC}\footnote{http://hasc.jp/hc2011/}~\shortcite{kawaguchi2011hasc,kawaguchi2011hasc2011corpus} is human activity data with 3 dimensions provided by HASC challenge 2011. We used the same subset of the HASC dataset from \emph{KL-CPD}~\shortcite{chang2019kernel}. 
    HASC dataset has type I sudden-change-points. 
\end{itemize}

\subsubsection{Synthetic Datasets}
\label{secData}
We generated two synthetic datasets to simulate different types of change-points in streaming data. Note that each data is piece-wise i.i.d. and has no time dependency within each segment or between segments. Both datasets have type I obvious-change and sudden-change-points.

\begin{figure}[!tb]
\begin{subfigure}{.9\textwidth}
  \centering
  \includegraphics[width=0.79\linewidth]{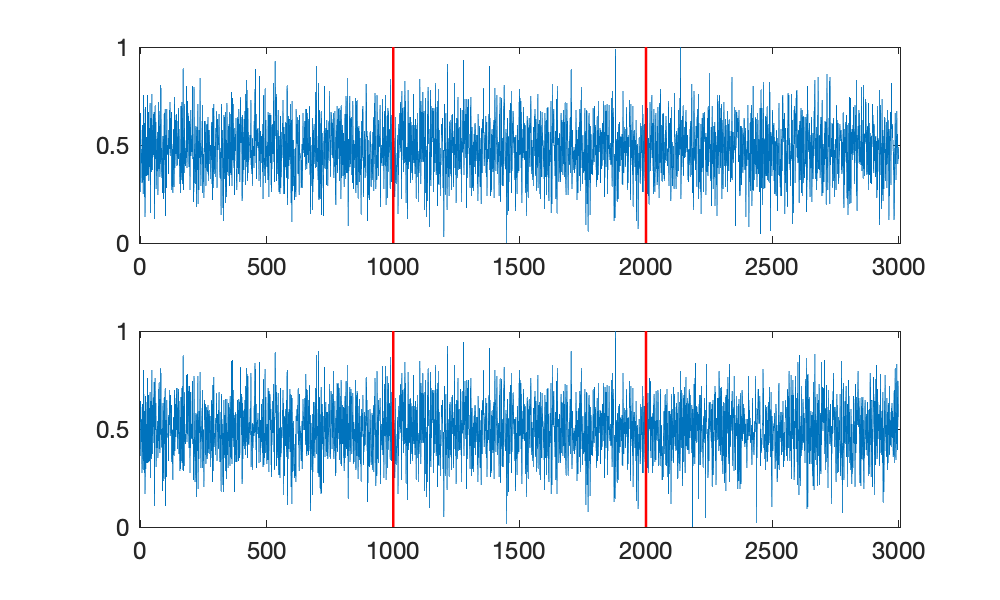}  
  \caption{1-dimensional plot}
  \label{fig:syn-2}
\end{subfigure}

\begin{subfigure}{0.9\textwidth}
  \centering
  \includegraphics[width=0.79\linewidth]{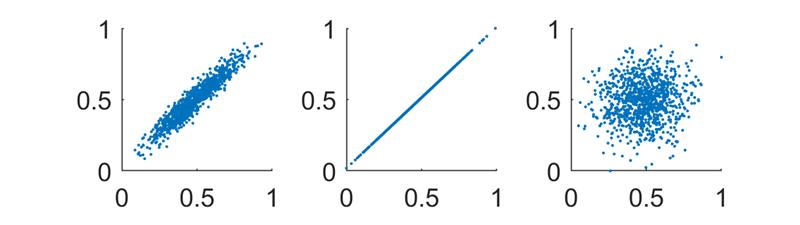}  
  \caption{2-dimensional plot}
  \label{fig:syn2-2}
\end{subfigure}
\caption{Data distribution of S2. (a) shows the data distribution on each attribute. (b) shows the data distribution using two attributes for each of the three intervals, split by the two change-points indicated as red bars in (a).}
\label{fig:syn2}
\end{figure}

\begin{itemize}
    \item \textbf{S1:} There are 5 one-dimensional Gaussians of 300 points with the same $\mu = 0$ but different $\sigma = 1, 2.2, 4.3, 48.3$, $28.3$, respectively. In addition, there are 5 outliers located at 89, 117, 139, 523 and 537, respectively. The data distribution is shown in Figure~\ref{fig:syn-1}.  
    \item \textbf{S2:} There are three two-dimensional Gaussians of 1000 points with the same $\mu = 0$ but different covariances $Cov=[0.9 \ 0.4;0.4 \ 0.2]$, $[0.5 \ 0.5;0.5 \ 0.5]$, $[0.9 \ 0.1;0.1 \ 0.9]$, 
    respectively. The data distribution is shown in Figure~\ref{fig:syn2}. Note that the data distribution on each attribute is the same over the whole period, while there are two type I change-points in the data when considering the two-dimensional covariance.  
\end{itemize}

\begin{table}[!tb]
	\renewcommand{\arraystretch}{1}
	\setlength{\tabcolsep}{5pt}
  \centering
  \caption{The properties of the datasets used in our experiments.}
    \begin{tabular}{c|ccc}
    \hline
    Dataset & Size & \#dimensionality &\#Changing interval \\
    \hline    
    S1 & 15000 & 1 & 4 \\
    S2 & 3000 & 2 & 2 \\ \hdashline
    USC-HAD & 93635 & 1 & 6 \\
    Yahoo-16 & 424 & 1 & 5 \\
    Google-trend & 219 & 1 & 2 \\
    Well\_log & 4050 & 1 & 10 \\
    Weather & 3288 & 1 & 2 \\
    HASC & 39397 & 3 &  65\\
    \hline
    \end{tabular}
  \label{tab:datasets}
\end{table}

\begin{figure}[!tb]
  \centering
\begin{subfigure}{.45\textwidth}
  \centering
  \includegraphics[width=0.99\linewidth]{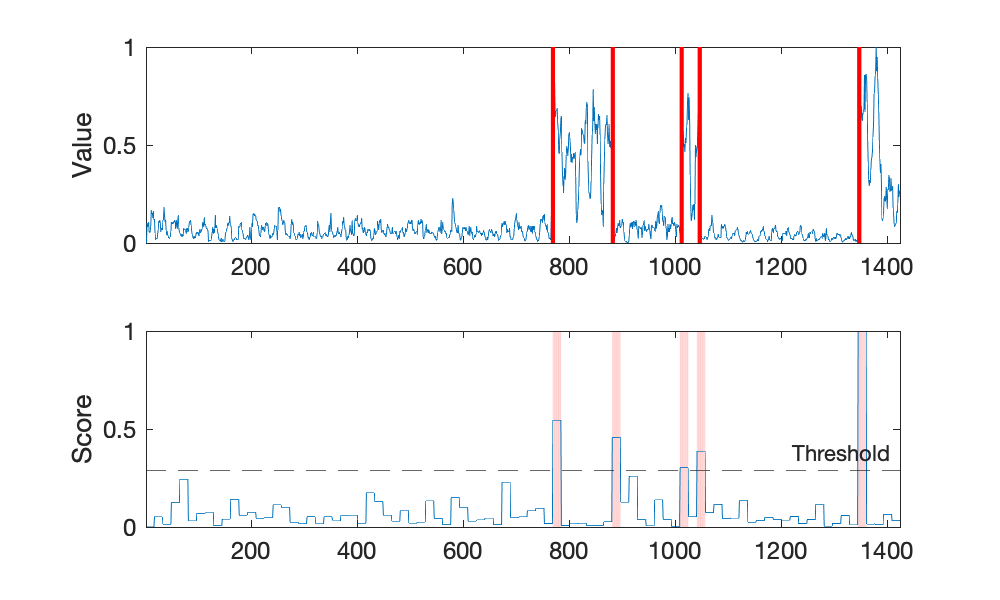}  
  \caption{Yahoo-16}
  \label{fig:offline-yahoo}
\end{subfigure}
\begin{subfigure}{.45\textwidth}
  \centering
  \includegraphics[width=0.99\linewidth]{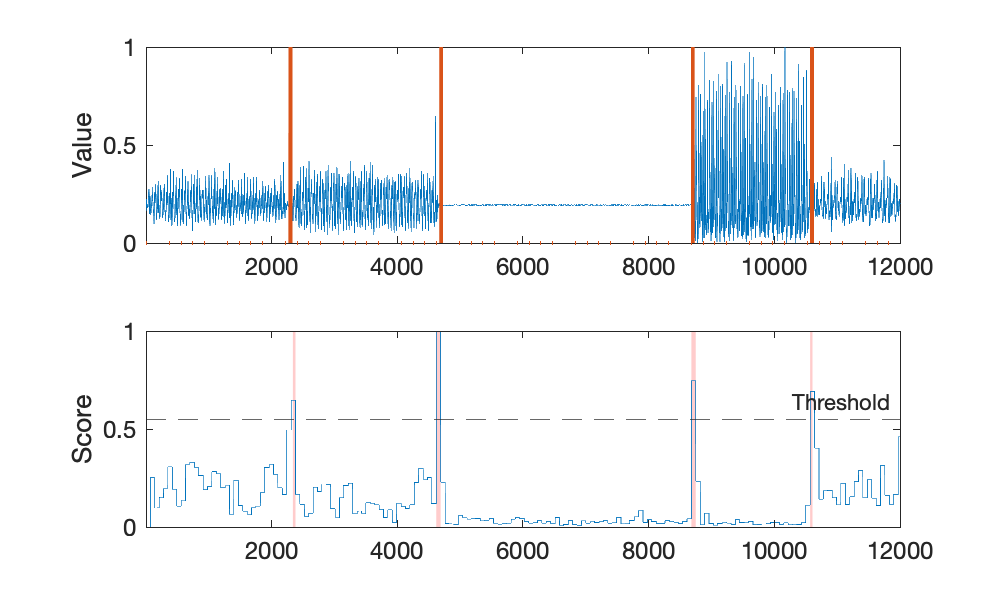}
  \caption{USC-HAD}
  \label{fig:offline-USC-HAD}
\end{subfigure}
\begin{subfigure}{.45\textwidth}
  \centering
  \includegraphics[width=0.99\linewidth]{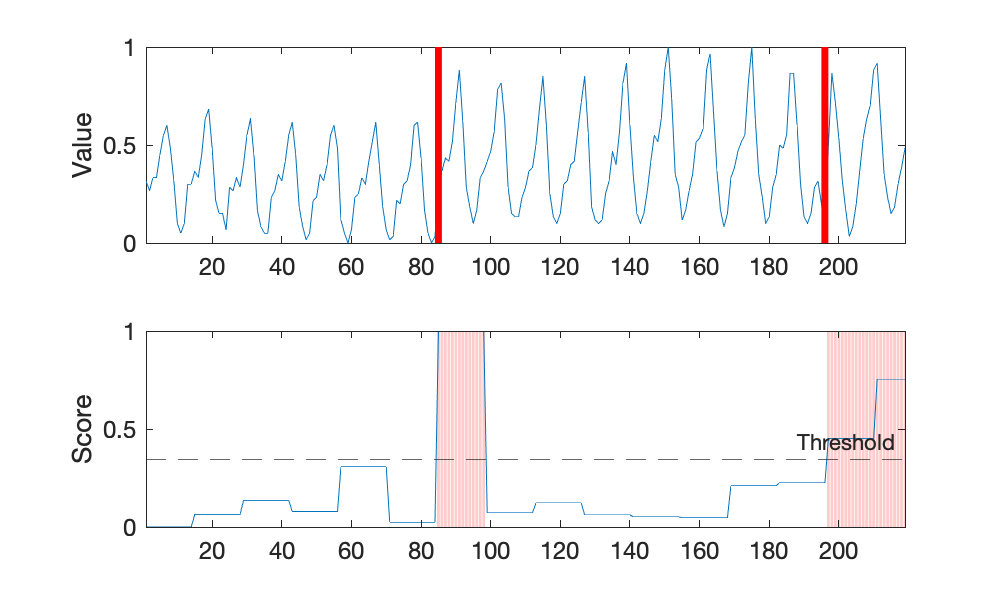}  
  \caption{Google-trend}
  \label{fig:offline-google}
\end{subfigure}
\begin{subfigure}{.45\textwidth}
  \centering
  \includegraphics[width=0.99\linewidth]{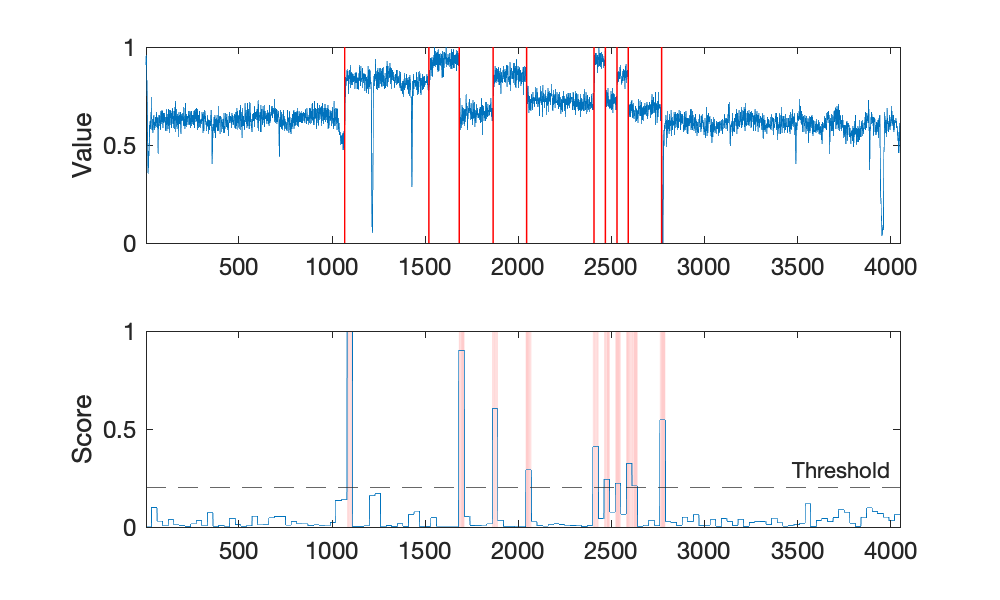}
  \caption{Well\_log}
  \label{fig:offline-well}
\end{subfigure}
\begin{subfigure}{.45\textwidth}
  \centering
  \includegraphics[width=0.99\linewidth]{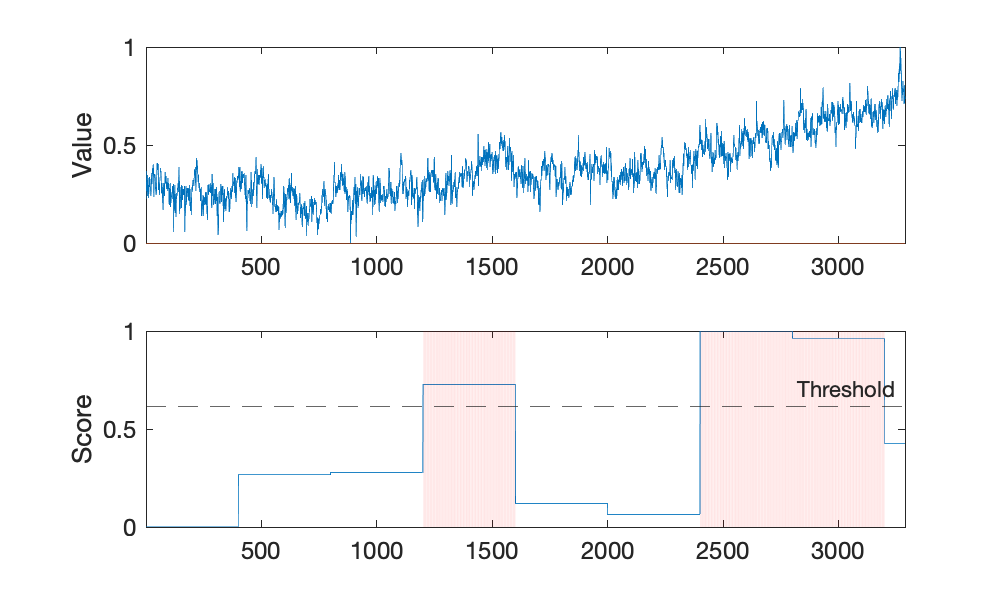}
  \caption{Weather}
  \label{fig:offline-weather}
\end{subfigure}
\begin{subfigure}{.45\textwidth}
  \centering
  \includegraphics[width=0.99\linewidth]{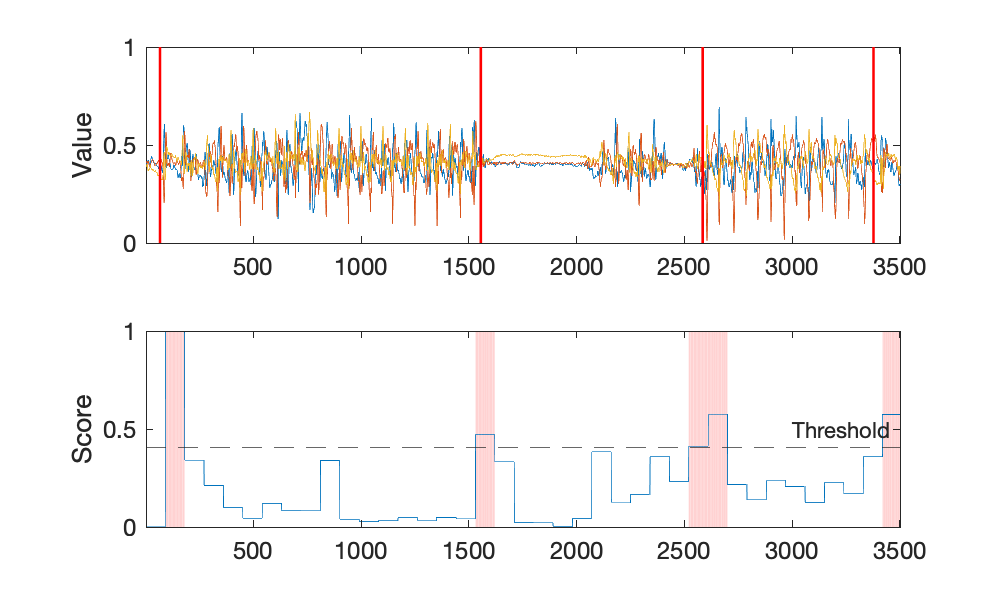}
  \caption{HASC}
  \label{fig:offline-hasc}
\end{subfigure}
\caption{Offline \texttt{iCID} results on real-world datasets. In each subfigure (on one dataset), the first row shows the data distribution with change-point locations marked by red bars. The second row shows the change scores and the detected change intervals, marked as red areas. Since the USC-HAD and HASC datasets are very large, here we only show the results on a segment that includes obvious changes for demonstration. }
\label{fig:offline}
\end{figure}

\subsection{Parameter Setting}

\begin{table}[!tb]
	\renewcommand{\arraystretch}{1}
	\setlength{\tabcolsep}{5pt}
	\centering
	\caption{Parameters search ranges for iCID}
	\begin{tabular}{l|l}
		\hline
		Parameter & Search ranges\\
		\hline

		 Subsampe  size & $\psi \in \{2, 4, 8, 16, 32, 64 \}$\\
		 Window size & $w\in \lceil 10, 15, ..., 400 \rceil$\\
          Power factor & $\alpha \in \lceil 0, 0.1, 0.2, ..., 3 \rceil$ \\
		\hline
	\end{tabular}
	\label{para}
\end{table}

Offline \texttt{iCID} has three inputs: Dataset $D$, window Size $w$ and Subsample Size List $\Psi$. For the online \texttt{iCID}, we used the first half of the whole datasets as the reference $D$ and applied a sliding window to generate the current streaming data $D'$. The size of sliding window was set to the half of the dataset length. The parameter search ranges are provided in Table \ref{para}.

\section{Empirical Evaluation}~\label{sec-result}

In this section, we aim to answer the following questions: 
\begin{itemize}
    \item Is \texttt{iCID} an effective method to identify three types of change-points in univariate and multivariate data streams? 
    \item Is \texttt{iCID} resistant to outliers and scalable to large data sizes?
    \item Does \texttt{iCID} have any advantage over other CPD algorithms, in terms of detection performance and time complexity?
\end{itemize}

To answer the first two questions, we present the empirical evaluation results  of the online and offline versions of \texttt{iCID} on the real-world and synthetic datasets in the next two subsections. The last two subsections provide the answer to the third question by  reporting the quantitative evaluation results on the two largest real-world datasets, in comparison with 6 state-of-the-art CPD algorithms.

\begin{figure}[!t]
    \centering
    \begin{subfigure}{.45\textwidth}
      \centering
      \includegraphics[width=0.99\linewidth]{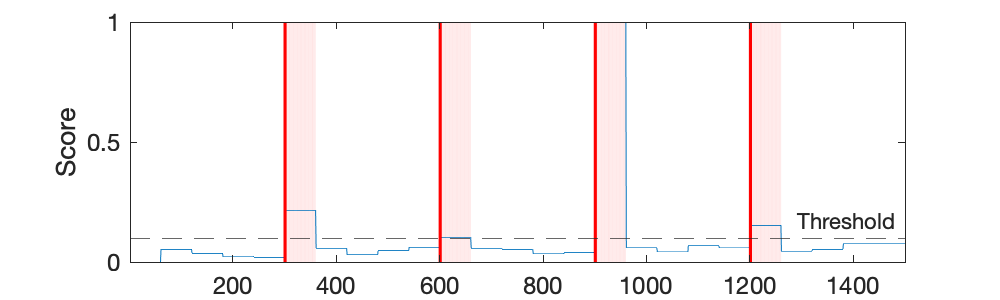}
      \caption{S1 dataset}
      \label{offline-syn1}
    \end{subfigure}
    \begin{subfigure}{.45\textwidth}
      \centering
      \includegraphics[width=0.99\linewidth]{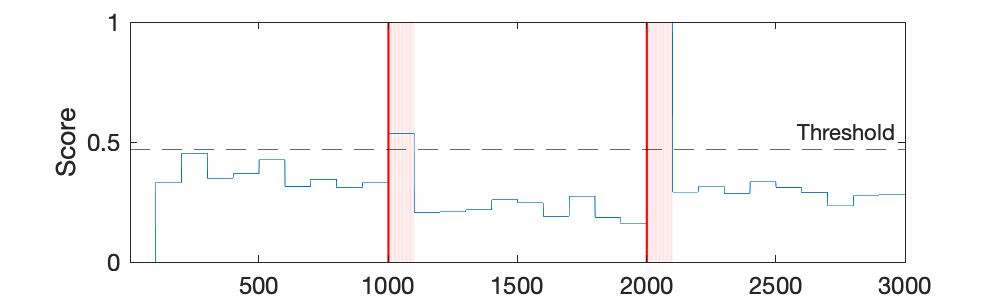}  
      \caption{S2 dataset}
      \label{fig:feature12}
    \end{subfigure}
    \begin{subfigure}{.45\textwidth}
      \centering
      \includegraphics[width=0.99\linewidth]{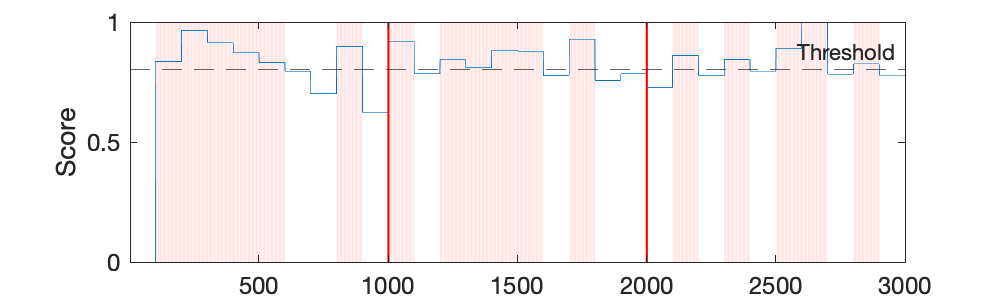}  
      \caption{S2 dataset (only using the 1st feature)}
      \label{fig:feature1}
    \end{subfigure}
\caption{Offline \texttt{iCID} results on synthetic datasets. (c) is the score of just using 2nd feature in S2 dataset.}
\label{offline-syn2}
\end{figure}

\subsection{Offline \texttt{iCID}}

The performance of offline \texttt{iCID} on 6 real-world datasets and 2 synthetic datasets are visualised in Figures~\ref{fig:offline} and \ref{offline-syn2}, respectively~\footnote{We also evaluated e-divisive and found it has shown good performance on S1 dataset, i.e., the identified top four change points are very close to the ground truth. We didn't run e-divisive on other datasets due to its high time complexity of $O(n^2)$.}. Note that a CID method detects a change interval that contains at least one ground truth change-points is regarded to have correctly identified a change interval, as stated in Definition \ref{def-change-interval}.

Observations on the real-world datasets are given below:

\begin{itemize}
    \item  In S2 dataset, there is no distribution change on any individual attribute. The change-points can only be detected by considering both attributes, as shown in  Figure~\ref{fig:syn2-2}.  The results presented in Figure~\ref{fig:feature12} show that \texttt{iCID} has the capability to detect the two change-intervals containing the change-points on this multivariate dataset.
    \item  The Google-trend dataset is a periodic dataset. 
    \texttt{iCID} is able to detect the change intervals which contain the all change-points.
    It is imperative to note that for optimal results in a periodic dataset, the window size should ideally approximate the length of a single period. 
    \item As shown in Figure~\ref{fig:offline-well}, the Well\_log dataset contains a few outliers which can easily be mistaken as change-points. 
    \texttt{iCID} successfully found 9 out of 10 change-points without misclassifying any outliers. 
    It is worth mentioning that all change-points can be correctly identified if manually set a large parameter $\psi$ value.
    \item The Weather dataset differs from other datasets 
    because it has both gradual (type II) and continuous (type III) change-points, as shown in Figure~\ref{fig:offline-weather}. 
    \texttt{iCID} discovers that there are two types of change-points in historical average temperatures, i.e., a short fluctuation around 1,500 and a continuously increasing trend after 2,500. 
\end{itemize}

In summary, the offline \texttt{iCID} detects all three types of change-points and can tolerate outliers with the effective and automatic parameter $\psi$ setting. It detected all change intervals that contain the change-points 
on all datasets. The only exception is one change-point on the Well\_log dataset.

\subsection{Online \texttt{iCID}}

\begin{figure}[!tb]
  \centering
\begin{subfigure}{.45\textwidth}
  \centering
  \includegraphics[width=0.99 \linewidth]{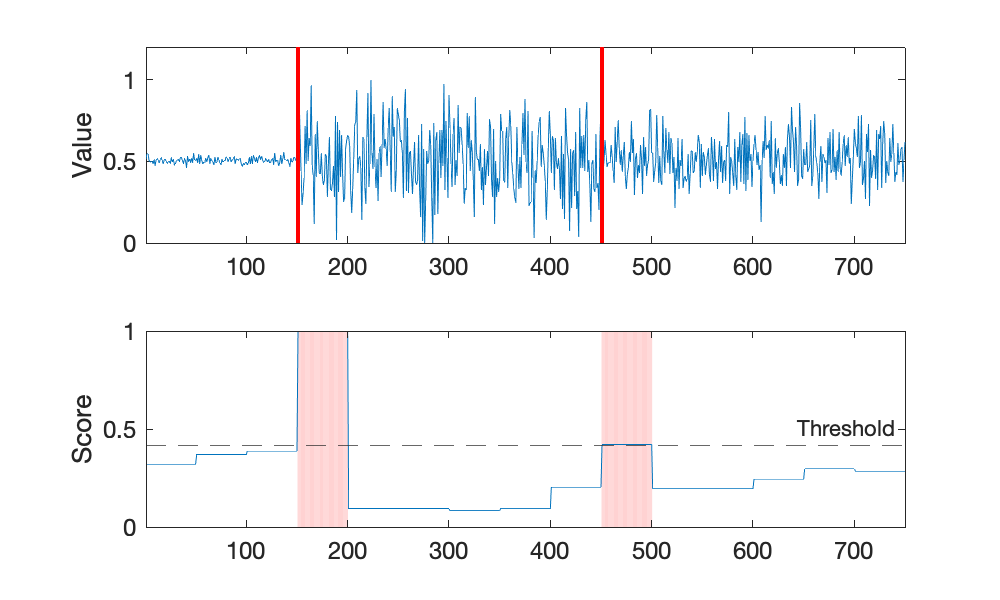}  
  \caption{S1}
  \label{online-syn1}
\end{subfigure}
\begin{subfigure}{.45\textwidth}
  \centering
  \includegraphics[width=0.99 \linewidth]{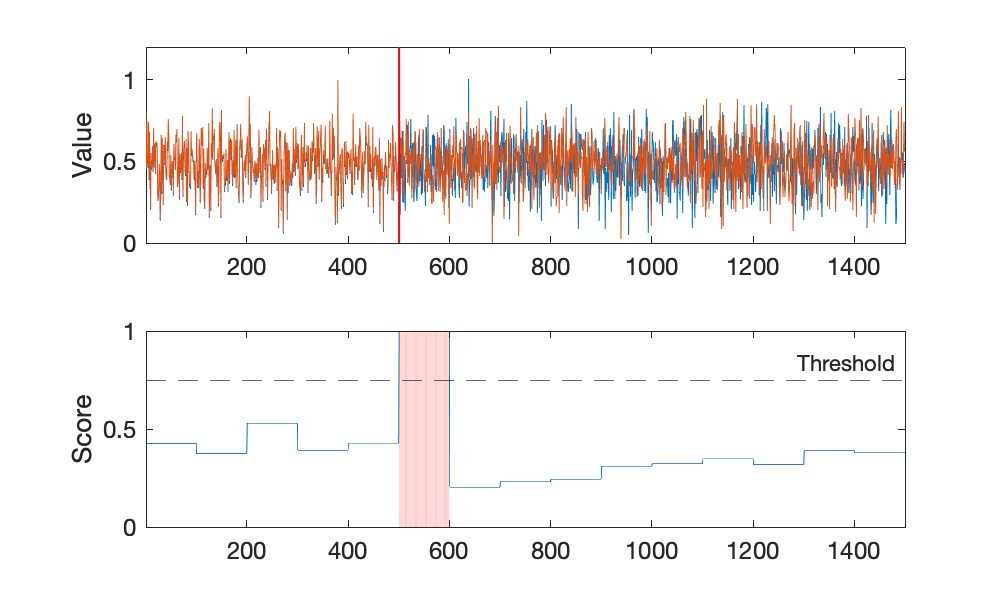}
  \caption{S2}
  \label{online-syn2}
\end{subfigure}
\caption{Online \texttt{iCID} results on the Synthetic datasets.}
\label{Ronline-syn}
\end{figure}

\begin{figure}[!t]
  \centering
\begin{subfigure}{.45\textwidth}
  \centering
  \includegraphics[width=0.99\linewidth]{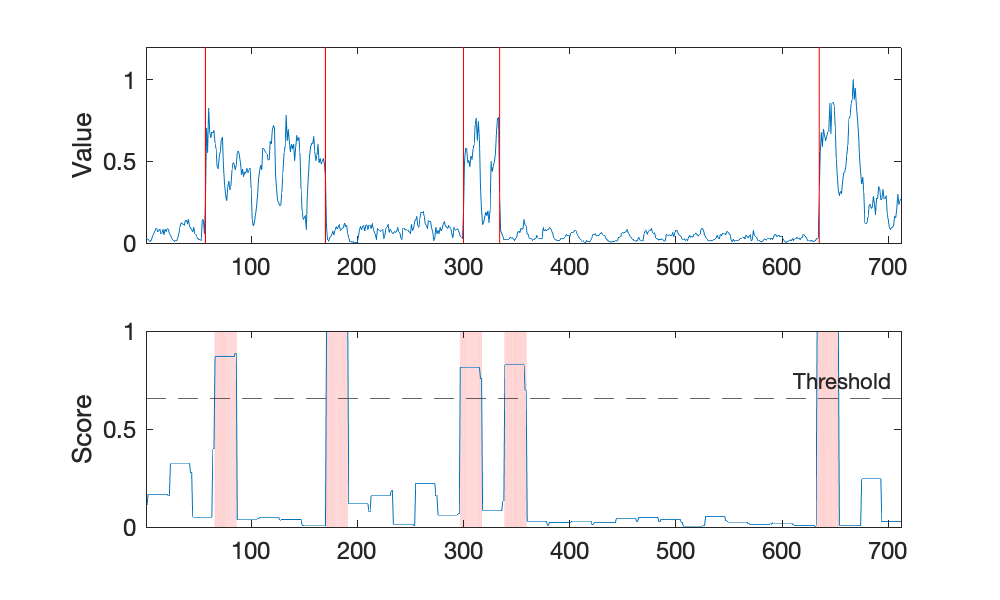}  
  \caption{Yahoo-16}
  \label{fig:online-yahoo}
\end{subfigure}
\begin{subfigure}{.45\textwidth}
  \centering
  \includegraphics[width=0.99\linewidth]{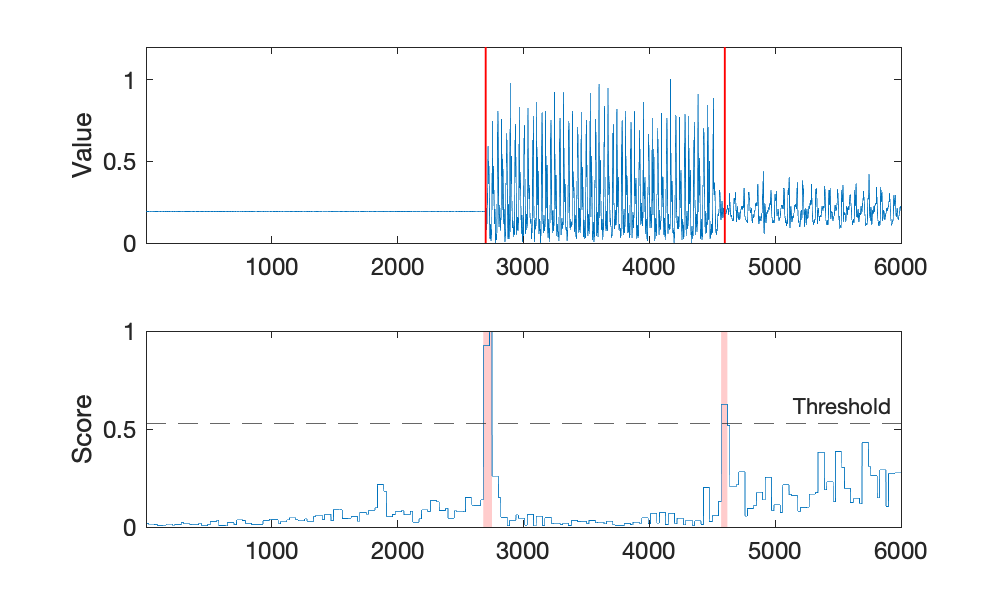}
  \caption{USC-HAD}
  \label{fig:online-USC-HAD}
\end{subfigure}
\begin{subfigure}{.45\textwidth}
  \centering 
  \includegraphics[width=0.99\linewidth]{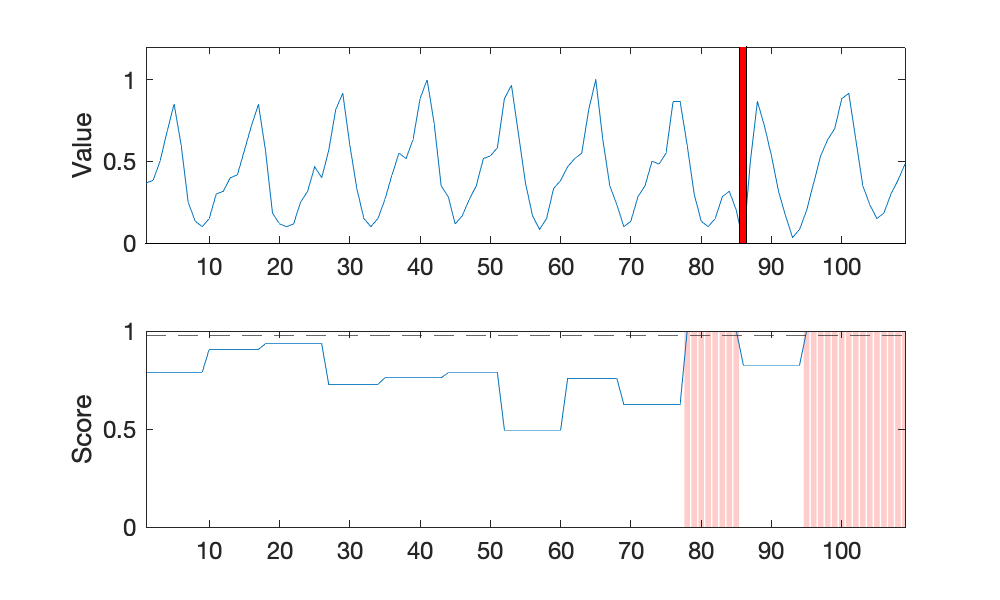}
  \caption{Google-trend}
  \label{fig:online-google}
\end{subfigure}
\begin{subfigure}{.45\textwidth}
  \centering 
  \includegraphics[width=0.99\linewidth]{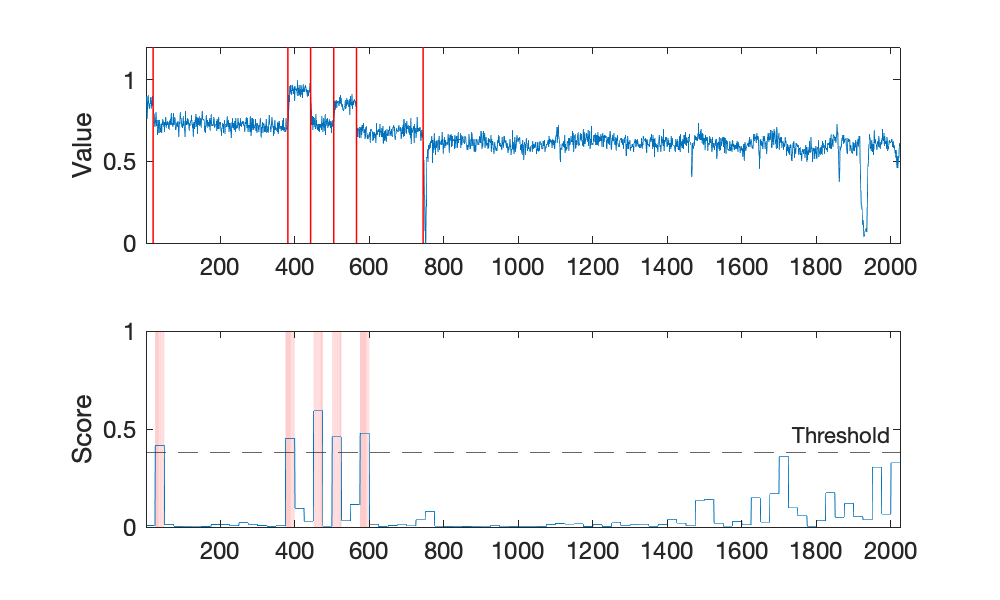}
  \caption{Well\_log}
  \label{fig:online-well}
\end{subfigure}
\begin{subfigure}{.45\textwidth}
  \centering 
  \includegraphics[width=0.99\linewidth]{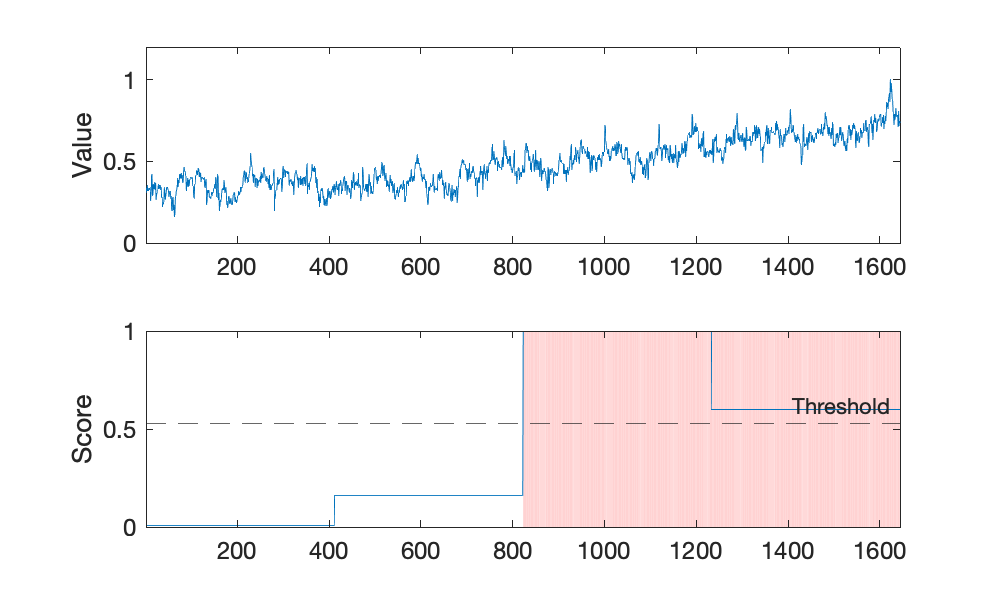}
  \caption{Weather}
  \label{fig:online-weather}
\end{subfigure}
\begin{subfigure}{.45\textwidth}
  \centering
  \includegraphics[width=0.99\linewidth]{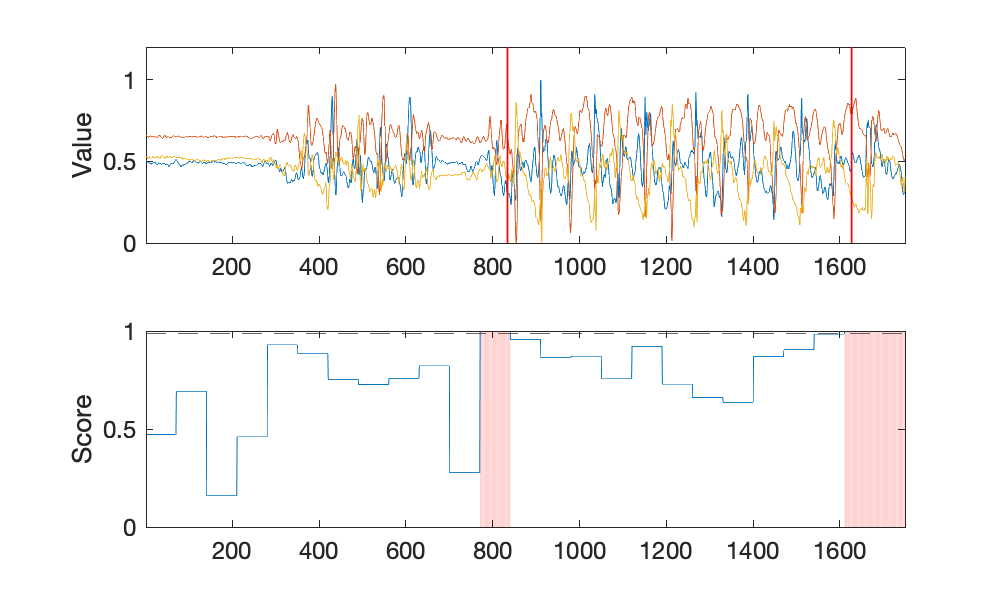}  
  \caption{HASC}
  \label{fig:online-hasc}
\end{subfigure}
\caption{Online CPD results on real-world datasets. In each subfigure, the first row is the data distribution with change-point location presented by red bars. The second row shows the change score and the detected change interval as red areas.}
\label{fig:online}
\end{figure}

The performance of online \texttt{iCID} evaluated on both synthetic datasets and  real-world  datasets are shown in Figures~\ref{Ronline-syn} and \ref{fig:online}, respectively.

We observe similar results to the offline version.
The online \texttt{iCID} can detect all change-points on all datasets except Well\_log dataset. 
Figure~\ref{fig:online-well} shows that \texttt{iCID} does not detect the 6th change-point on the Well\_log dataset, but this change-point should be considered as an anomaly instead of a change-point according to Definition~\ref{changePoint}. 
Moreover, online \texttt{iCID} produces a false-positive change-point on the Google-trend dataset as shown in Figure~\ref{fig:online-google}. This error can be easily rectified by manually setting a better $\psi$ value. 

It is worth mentioning that the performance online version is a bit lower than that of the offline version. Recall that online \texttt{iCID} requires $\mathcal{D}$ used in IDK to be continuously updated from a limited number of previous points. As a result, $\mathcal{D}$ in online \texttt{iCID} may be less representative than it sampled from the whole dataset used in the offline \texttt{iCID}. 

\begin{table}[!tb]
	\renewcommand{\arraystretch}{1}
	\setlength{\tabcolsep}{5pt}
\centering
  \caption{Performance comparison with state-of-the-art methods on the two largest real-world datasets. The highest F1-score on each dataset with the same detection margin is boldfaced. The results of existing methods are from the \emph{TS-CP$^{2}$} paper \shortcite{deldari2021time}.} 
  \label{tbl:experiments}
  \resizebox{\linewidth}{!}{
  \begin{tabular}{c|c|cc|cc|cc}
\hline
Dataset & Methods & Best Window & F1-score & Best Window & F1-score & Best Window & F1-score\\
\hline
\multirow{11}*{HASC} & Detection Margin & \multicolumn{2}{c|}{60} & \multicolumn{2}{c|}{100} &  \multicolumn{2}{c}{200} \\
\hline
~                   & \emph{aHSIC}            &  40     & 0.2308       &  40         &  0.3134   &  40   &   0.4167        \\
~                   & \emph{RuLSIF}           &  200     & 0.3433       &  200         &  0.4999   &  200   &   0.4999        \\
~                   & \emph{ESPRESSO}         &  100     & 0.2879       &  60         &  0.4233   &  100   &   0.6933        \\
~                   & \emph{Mstat}         &  35     & 0.2958       &  35         &  0.4246   &  15   &   0.6372      \\
~                   & \emph{KL-CPD}           &  60     & \textbf{0.4785}       &  100         &  0.4726   &  200   &   0.4669        \\
~                   & \emph{TS-CP$^{2}$}        &  60     & 0.40       &  100         &  0.4375   &  200   &   0.6316        \\
\cline{2-8}
~                   & $\texttt{gCID(MMD)}$      &  65    &   0.3522     &  55      &   0.5493    &  90   &  0.7813      \\
~                   & $\texttt{iCID(MMD)}$     &  40     & 0.3117       &  50         &  0.4734   &  100   &   0.7612 \\
~                   & $\texttt{iCID}$      &  65     & 0.3333       &  85         &  \textbf{0.5630}   &  120   &   \textbf{0.7943}        \\ 
\hline
\hline
\multirow{11}*{USC-HAD} & Detection Margin & \multicolumn{2}{c|}{100} & \multicolumn{2}{c|}{200} &  \multicolumn{2}{c}{400}\\ \hline
    &   \emph{aHSIC} & 50    & 0.3333 & 50    & 0.3333 & 50    & 0.3999 \\
    &   \emph{RuLSIF} & 400   & 0.4666 & 400   & 0.4666 & 400   & 0.5333 \\
    &   \emph{ESPRESSO} & 100   & 0.6333 & 100   & 0.8333 & 100   & 0.8333 \\
~                   & \emph{Mstat}         &  35     & 0.5185       &  30      &  0.5352   &  30   &   0.6774     \\ 
  & \emph{KL-CPD} & 100 & 0.7426 & 200 & 0.7180 & 400 & 0.6321\\
  &   \emph{TS-CP$^{2}$}  & 100 & \textbf{0.8235} & 200 & \textbf{0.8571} & 400 & 0.8333\\
\cline{2-8}
~                   & $\texttt{gCID(MMD)}$      &  60    &  0.6786      &   65     &  0.7941     &  255   & \textbf{0.8451}       \\ 
~ 
& $\texttt{iCID(MMD)}$      &  70    &  0.7273      &   65     &  0.7632     &  255   & 0.8421       \\
~ 
& $\texttt{iCID}$      &    70    &    0.6857    &    70      & 0.7536     &   255  &     {0.8378}     \\
\hline
\end{tabular}}
\end{table}

\subsection{Comparison With State-of-the-Art Methods}
In this subsection, we compare the performance of $\texttt{iCID}$ against 6 state-of-the-art (SOTA) methods, including \emph{ESPRESSO}~\shortcite{deldari2020espresso},  \emph{aHSIC}~\shortcite{yamada2013change}, \emph{RuLSIF}~\shortcite{liu2013change},  \emph{KL-CPD}~\shortcite{chang2019kernel} and \emph{TS-CP$^{2}$}~\shortcite{deldari2021time}, on the two largest real-world datasets, i.e., USC-HAD and HASC. 

For a fair comparison, we followed the same evaluation strategy used previously \shortcite{deldari2021time} and calculated the highest F1-score of $\texttt{iCID}$ with manually tuned parameters and detection margins. Only if a detected change-point/interval is located within the specified margin of the ground truth change-point, then it is a true positive~\footnote{Researchers can also expand the change points detected by existing change point methods to change intervals of the same width as used in proposed method, and then evaluate as done for the proposed method.}.

Table~\ref{tbl:experiments} presents the performance comparison.  We have the following observations regarding the comparison between $\texttt{iCID}$ and the 6 SOTA algorithms:
\begin{itemize}
    \item $\texttt{iCID}$ has the highest F1 score on the HASC dataset when the detection margin is 100 and 200. The best window sizes are only around half of the corresponding detection margins. 
    \item \emph{ESPRESSO} is a shape-based CPD method, its performance is usually far below than that of $\texttt{iCID}$ on both datasets under different margin values. The reason can be the fact that shape-based CPD methods are intolerance to outliers.
    \item Comparing the three Gaussian-based kernel methods, i.e., \emph{Mstat},  \emph{aHSIC} and  \emph{RuLSIF}, $\texttt{iCID}$ outperforms them on both datasets with the detection margin greater than or equal to 100. 
    \item  With a large detection margin, $\texttt{iCID}$ performs significantly better than \emph{TS-CP$^{2}$} on HASC, but is comparable to \emph{TS-CP$^{2}$} on USC-HAD. 
    \emph{TS-CP$^{2}$} is a self-supervised deep learning-based method that needs heavy training, whereas $\texttt{iCID}$ requires no learning. 
    \item The F1 score of a model shall increase if the detection margin increases, i.e., a larger margin reduces potential misalignments between the labels and the change score due to lag. 
    However, the F1 scores of two deep learning-based methods, i.e., \emph{KL-CPD} and \emph{TS-CP$^{2}$}, decrease as the detection margin increases. The reason can be the instability of these deep learning methods.
    \item Based on the results on the largest detection margin, \texttt{iCID}'s closest contenders are \emph{TS-CP$^{2}$} and \emph{ESPRESSO} on both datasets. \texttt{iCID} win on a large F1 difference on HASC and a small one on USC-HAD. 
\end{itemize}

We also investigate two variants of $\texttt{iCID}$, i.e., $\texttt{iCID(MMD)}$ and $\texttt{gCID(MMD)}$, both use \emph{MMD} \shortcite{MMD-2006}. The latter uses Gaussian distributional kernel in MMD, as used in the two-sample test algorithms \emph{Mstat} and \emph{KL-CPD}; and the former uses Isolation distributional kernel as used in $\texttt{iCID}$.

It is interesting to mention that $\texttt{iCID}$  and its two variants show similar performance  on the datasets, as shown in Table \ref{tbl:experiments}. Moreover, $\texttt{gCID(MMD)}$ shows much better performance than \emph{Mstat} and  \emph{KL-CPD}.  
The reason can be using a fixed number of reference blocks before the potential change-point used in \emph{Mstat}. Some of those blocks may be outdated. Since the parameter of \emph{KL-CPD} is searched based on deep learning, it may not be optimised to detect different types of change-points.

\subsection{Scalability Evaluation}

\begin{figure}[!tb]
    \centering
    \includegraphics[width = 0.55\textwidth]{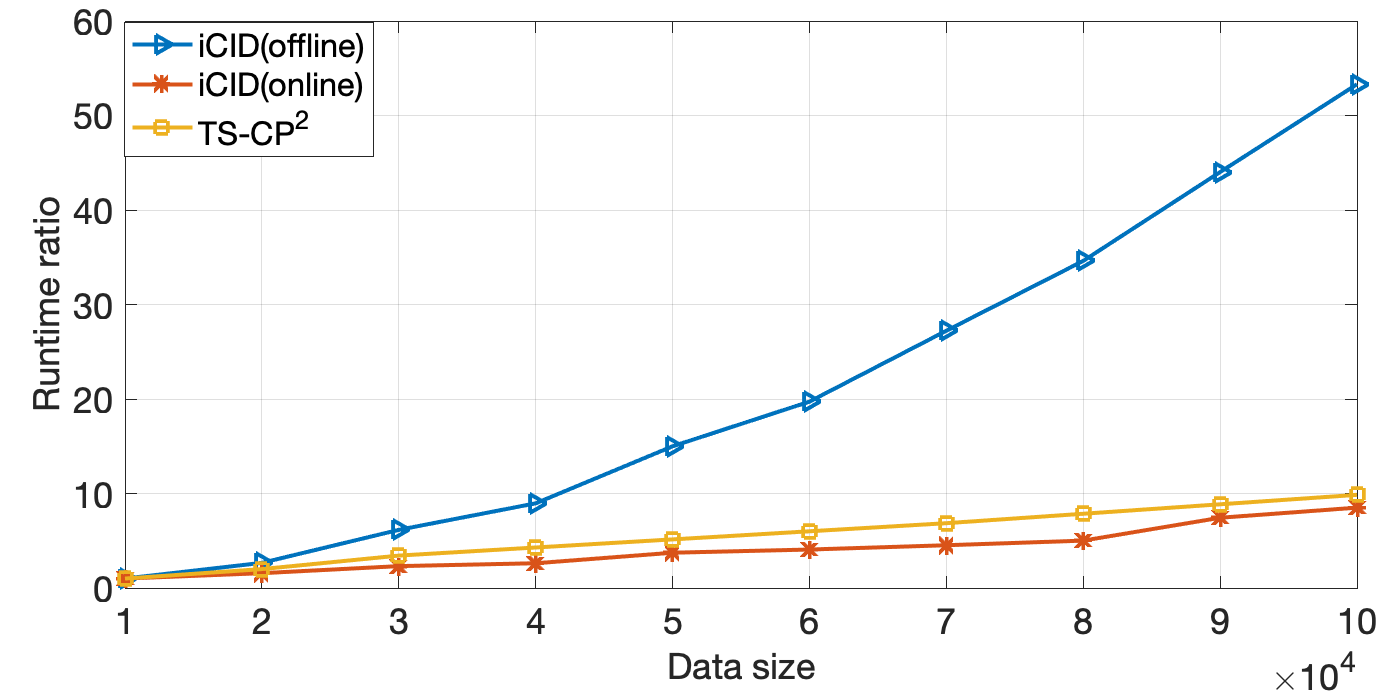}
    \caption{Scaleup test on a synthetic dataset. 
    \texttt{iCID} is tested using Matlab R2022b with Intel(R) Xeon(R) Gold 6154 CPU @ 3.00GHz and \emph{TS-CP$^{2}$} is tested using Python with NVIDIA Quadro RTX 5000 GPU. For \emph{TS-CP$^{2}$}, We include the total time of both training and testing processes. For \texttt{iCID}, we include the parameter search and model update time for offline and online versions, respectively.} 
    \label{fig:time}
\end{figure}

The runtime ratios of \texttt{iCID} and \emph{TS-CP$^{2}$} against a different number of points are shown in Figure~\ref{fig:time}. The results show that both versions of \texttt{iCID} have linear time complexity. Note that the offline \texttt{iCID} needs to run \texttt{iCID} multiple times and measure the stability in order to determine the best $\psi$ value, but online \texttt{iCID} does not need to search the parameters. Thus, the runtime ratio of offline \texttt{iCID} is higher than that of online \texttt{iCID}. 
 
It is interesting to point out that the exact CPU runtimes for the offline and online versions of \texttt{iCID}  on a data stream with 100,000 points are 203 seconds and 4 seconds, respectively. Yet,  a deep learning-based method \emph{TS-CP$^{2}$} took about 170 GPU hours to train with the default parameter setting using a high-end GPU. Although \emph{TS-CP$^{2}$} shows a linear runtime due to using a fixed number of epochs, it will cost much more time when the best parameters are to be tuned. Note that although the online \emph{TS-CP$^{2}$} also runs a few seconds once the model has been trained, it still needs a long time to retrain and update the model to adapt to a new data distribution.

\subsection*{Section Summary}
We find that iCID has three advantages over existing methods. 
\begin{enumerate}
    \item  iCID can handle data with subtle-change-points and perform better than data-independent kernel-based methods such as \emph{Mstat} which is shown in Figure~\ref{fig:syn1}. 
    \item  iCID has a linear time complexity and does not require training. As a result, it runs much faster than deep learning-based methods in real-time. 
    \item In the online setting, iCID can select the best parameter to adapt to a new data distribution in real time; but deep learning-based methods have difficulty retraining or updating  a model online in a short time. 
\end{enumerate}

\section{Discussion}~\label{sec-discussion}

In this section, we first present the change-point detection task based on \texttt{iCID}, and then discuss the label issues in the real-world datasets as mentioned in Section~\ref{cid}. Moreover, we conduct a sensitivity analysis of \texttt{iCID} with respect to different parameter settings.

\subsection{Change-Point Detection with IDK}
\label{sectionCPD}

\begin{figure}[!tb]
    \centering
    \includegraphics[width =0.54\textwidth]{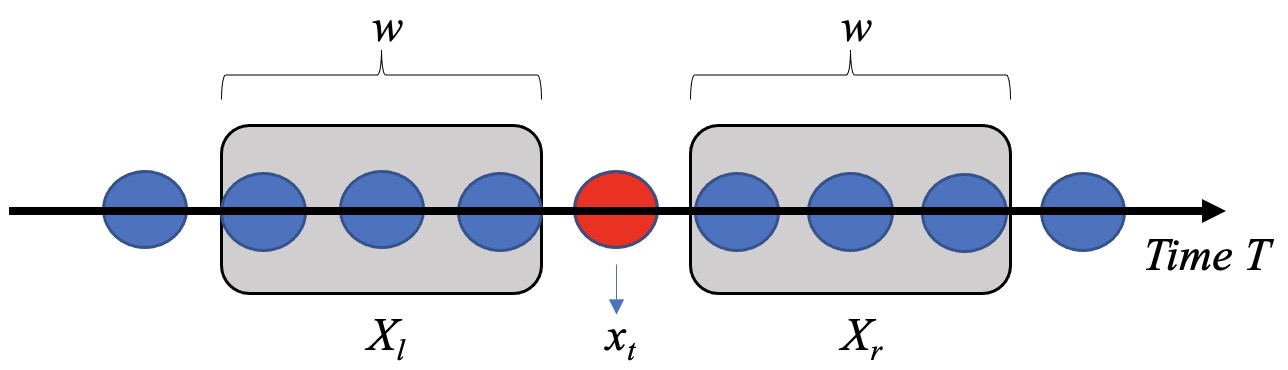}
    \caption{Overview of change-point detection based on IDK. }
    \label{fig:CPD}
\end{figure}

Using IDK for change-point detection is a special case of change-interval detection. Let the past interval $X^t_l=\{x_{t-w}, x_{t-w+1},... , x_{t-1}\}$ and the current interval $X^t_r=\{x_{t+1}, x_{t+2},... , x_{t+w}\}$ at each point $x_t$, as shown in Figure \ref{fig:CPD}. The change-point is the $x_t$ which has a large score more than a threshold $\tau$, i.e., $\mathfrak{S}(X^t_l, X^t_{r}) \geq \tau$. 

In order to detect all change-points in a data stream, we only need to slide the $w$-size intervals by one point to score every point in the stream. As a result, this \texttt{iCPD} method  needs significantly more computational time than the interval-based \texttt{iCID}. This is because \texttt{iCID} computes a score for each non-overlapping interval only. The time complexity is the same to \texttt{iCID} with $w$ times slower than \texttt{iCID}.
Figure \ref{fig:syn1-idk} shows the result of iCID using the sliding window method on the S1 dataset.

\subsection{About Ground-Truth Label Issues}
\begin{figure}[!tb]
    \centering
    \includegraphics[width =0.7\textwidth]{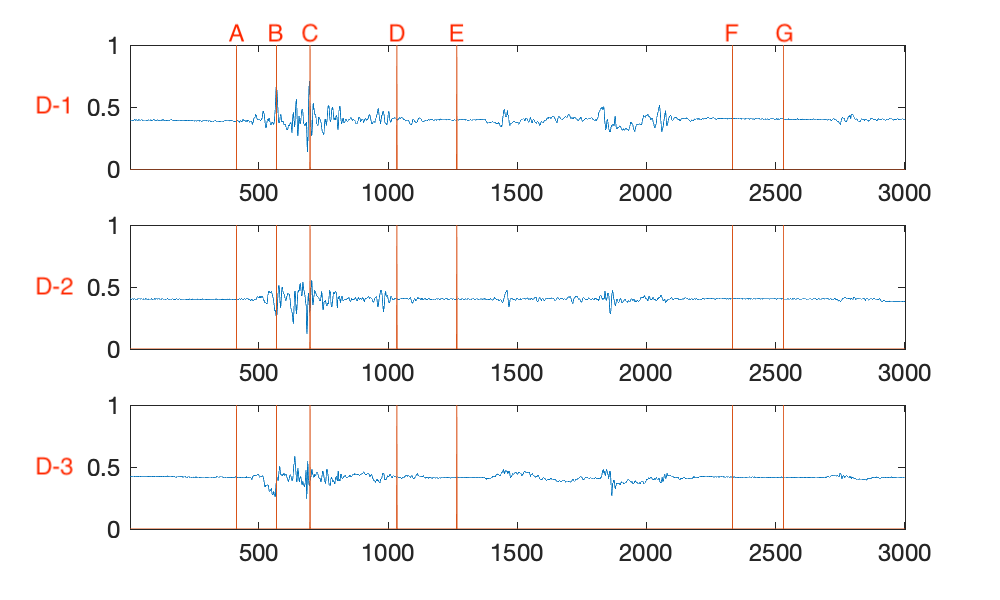}
    \caption{Demonstration of label issues on HASC dataset. D-1, D-2 and D-3 refer to different features.}
    \label{fig:issue}
\end{figure}

We visualised our evaluation of real-world datasets and found that many of them have different labelling problems. 
To illustrate, Figure~\ref{fig:issue} presents an interval of the 3-dimensional HASC dataset. Here we observe three types of issues as follows:

\begin{itemize}
    \item \textbf{Wrong label position:} change-points are labelled at incorrect or inaccurate locations. For label D, the location of the label is far away from the exact change-point. 
    \item \textbf{No-labelling:} change-points are not labelled. Around point 2000, the distribution of the data has some significant change but none of the points is labelled as change.
    \item \textbf{Mislabelling:} normal points or outliers are mistaken for change-points. For labels A, E, F and G, we cannot find any distribution change before and after these points in all three dimensions.
\end{itemize}

Similar labelling issues have been identified in time series datasets~\shortcite{wu2021current,freeman2021experimental}.
Therefore, due to the presence of error labels, Table~\ref{tbl:experiments} cannot accurately represent the best performance of different CPD algorithms. However, even with the existence of label errors in the datasets, our proposed algorithm \texttt{iCID} still achieves the state-of-the-art performance in F1 score.

\subsection{Multi-Dimensional Evaluation}

In Section~\ref{sec-result}, we evaluated the performance of \texttt{iCID} on two multi-dimensional datasets: S2 (2-dimensional) and HASC (3-dimensional). To test the scalability of iCID on high-dimensional data, we conducted an additional experiment on the MNIST dataset, which has 784 dimensions and 10 classes. We simulated a data stream by ordering the appearance of each class one after another.

\begin{figure}[!tb]
    \centering
    \includegraphics[width=0.7\linewidth]{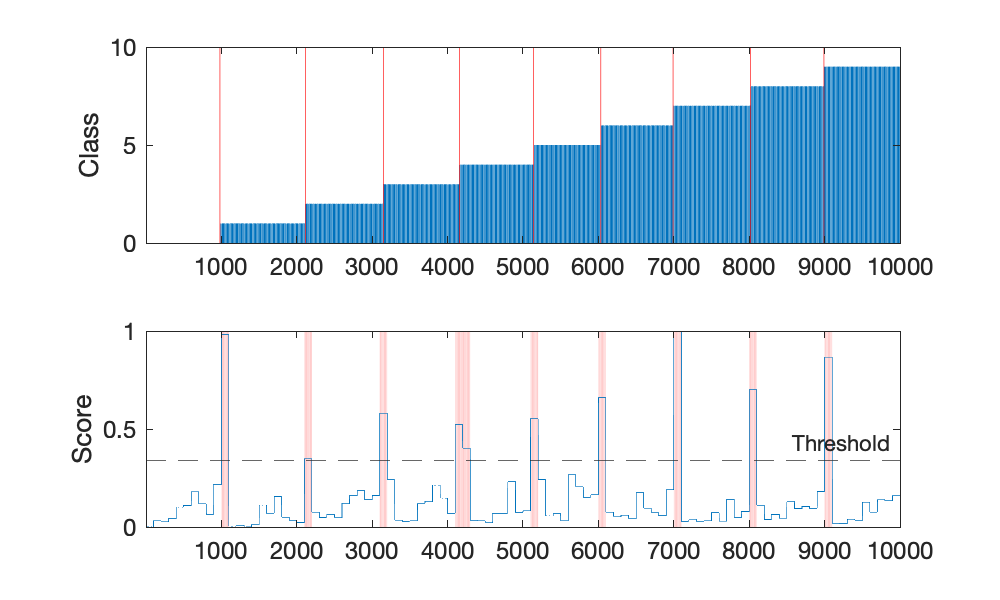}
    \caption{Offline \texttt{iCID} result on MNIST datasets. The first figure shows the 10 ordered classes in the MNIST dataset with change-point locations indicated by red bars. The second row shows the change score and the detected change intervals are marked as red areas.}
    \label{fig:high}
\end{figure}

Figure~\ref{fig:high} shows that
the offline \texttt{iCID} can successfully detect all the change intervals. This implies that \texttt{iCID} can perform well if the distributional change between adjacent intervals can be measured using the Isolation distributional kernel in multi-dimensional datasets. 

Having said that, when a high-dimensional dataset is plagued with the curse of high dimensionality which has multi-faceted issues (see e.g., \shortcite{Keogh2017} ), the true distributional change can be masked. In this case, no methods can do well.

\subsection{Parameter Sensitivity Analysis}
\label{sensitest}

There are three parameters in the offline \texttt{iCID}, i.e., the subsample size $\psi$ used to build Isolation distributional kernel, the interval/window size $w$ and the power factor $\alpha$ used to set the threshold. 

\begin{figure}[!htbp]
  \centering
\begin{subfigure}{.48\textwidth}
  \centering
  \includegraphics[width=0.99\linewidth]{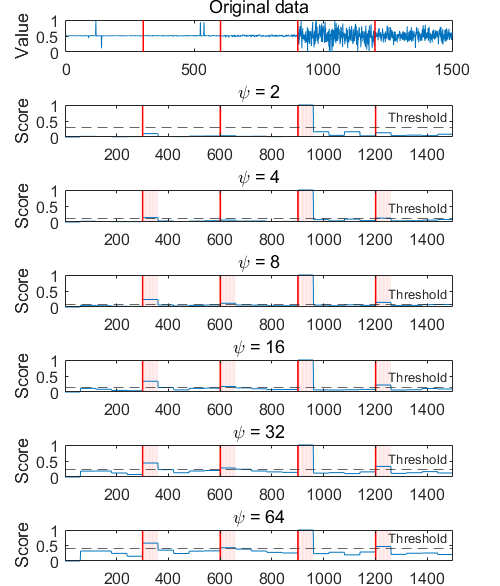}  
  \caption{S1 dataset with $w = 60$}
  \label{fig:sa-syn1}
\end{subfigure}
\begin{subfigure}{.48\textwidth}
  \centering
  \includegraphics[width=0.99\linewidth]{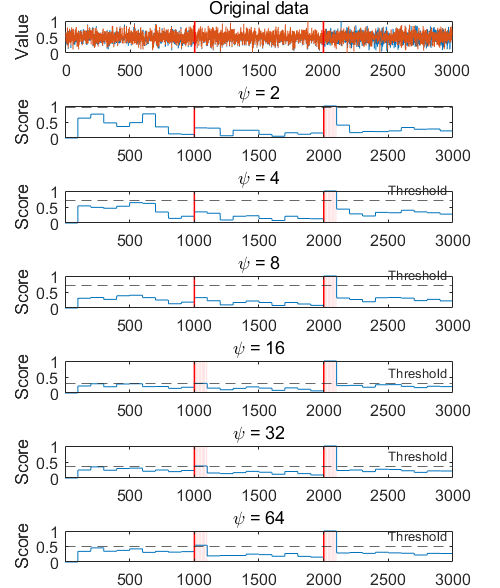}
  \caption{S2 dataset with $w = 100$}
  \label{fig:sa-syn2}
\end{subfigure}
\begin{subfigure}{.48\textwidth}
  \centering
  \includegraphics[width=0.99\linewidth]{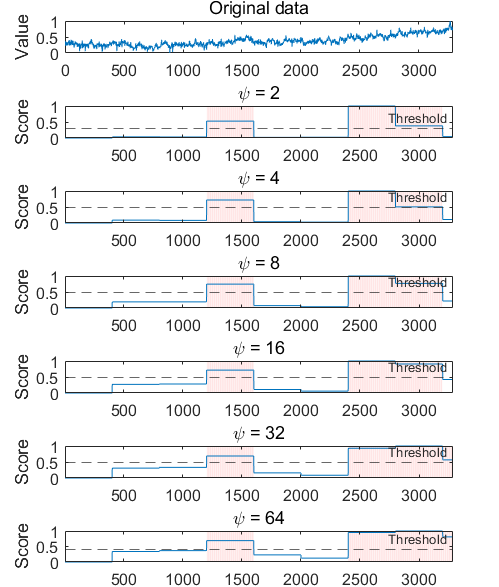}  
  \caption{Weather dataset with $w = 400$}
  \label{fig:sa-weather}
\end{subfigure}
\begin{subfigure}{.48\textwidth}
  \centering
  \includegraphics[width=0.99\linewidth]{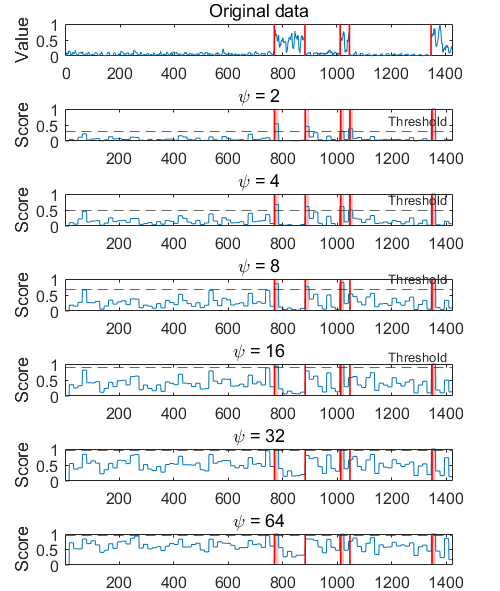}  
  \caption{Yahoo-16 dataset with $w = 16$}
  \label{fig:sa-yahoo}
\end{subfigure}
\caption{Sensitivity analysis of $\psi$ on four datasets.}
\label{fig:sa}
\end{figure}

Generally, a larger $\psi$ will make a sharper dissimilarity distribution, similar to a smaller bandwidth set in Gaussian kernel. For unsupervised learning, we have shown that $\psi$ determined by Equation (\ref{eqn:psi}) can produce promising performance in the previous section. Figure \ref{fig:sa} illustrates the effects of different $\psi$ values on four datasets. It can be seen from the results that a large $\psi$ value can perform well on these datasets. It is worth mentioning that $\psi^*$ searched by our proposed method is close to the best $\psi$ value on most datasets.

Regarding interval length $w$ setting for CID methods in practice, a smaller length of the interval is preferable to detect the change interval sooner than using a larger length of the interval.

\begin{figure}[!htbp] 
\centering
\begin{subfigure}{.48\textwidth}
  \centering
  \includegraphics[width=0.99\linewidth]{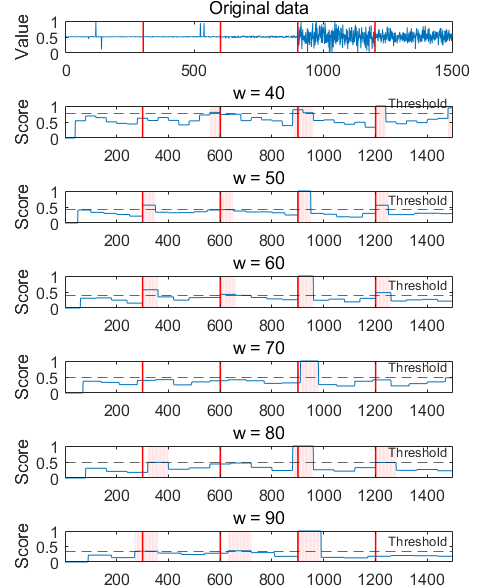}
  \caption{S1  dataset with $\psi = 64$}
  \label{fig:saw-syn1}
\end{subfigure}
\begin{subfigure}{.48\textwidth}
  \centering
  \includegraphics[width=0.99\linewidth]{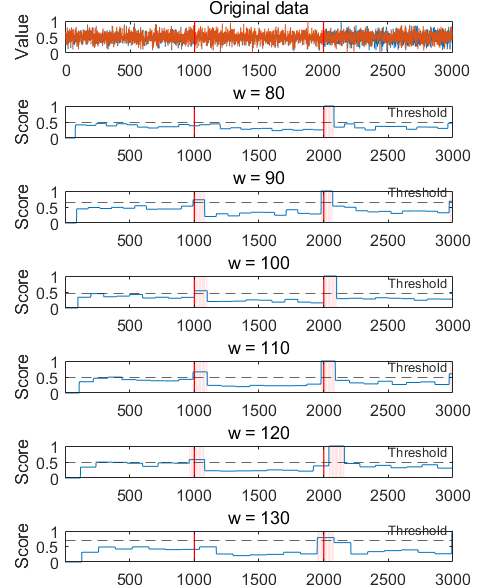}
  \caption{S2  dataset with $\psi = 64$}
  \label{fig:saw-syn2}
\end{subfigure}
\begin{subfigure}{.48\textwidth}
  \centering
  \includegraphics[width=0.99\linewidth]{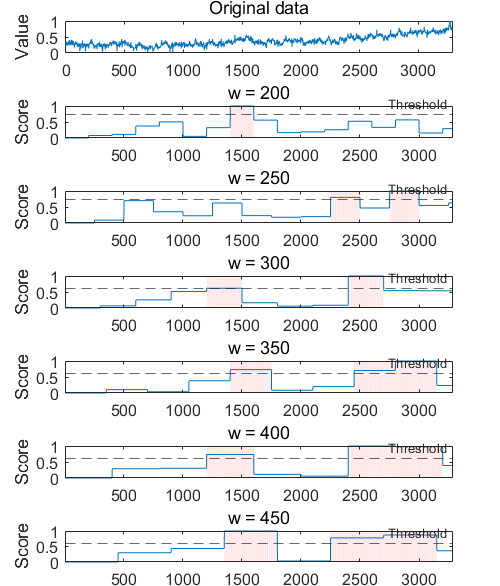}
  \caption{Weather dataset with $\psi = 16$}
  \label{fig:saw-weather}
\end{subfigure}
\begin{subfigure}{.48\textwidth}
  \centering
  \includegraphics[width=0.99\linewidth]{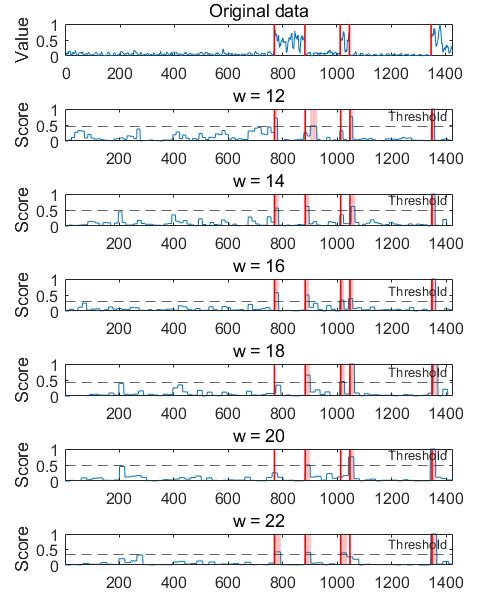}
  \caption{Yahoo-16 dataset with $\psi = 2$}
  \label{fig:saw-yahoo}
\end{subfigure}
\caption{Sensitivity analysis of $w$ on four datasets.}
\label{fig:sa2}
\end{figure}

Figure \ref{fig:sa2} shows the sensitivity analysis of $w$ on four datasets. The results indicate that the best $w$ is varied over different datasets. Although the window size $w$ should contain a sufficient amount of points to represent the data distribution,  a smaller length is preferable to detect the change of distribution sooner. 

The other parameter $\alpha$  in Equation~(\ref{eqn:threshold}) is a power factor parameter that denotes the level of sensitivity for change interval identification. When we assume that the distribution of change scores conforms or is close to a Gaussian distribution, $\alpha$ can control the expected proportions of the sample as change intervals, e.g. any point which is larger than $\mu + 3 \sigma$ will be considered as a change-point when $\alpha = 3$.  
In practice, we can set $\alpha\in [1,3]$ and a higher $\alpha$ will filter more subtle change-points.

\begin{figure}[!htbp]
  \centering
\begin{subfigure}{.48\textwidth}
  \centering
  \includegraphics[width=0.99\linewidth]{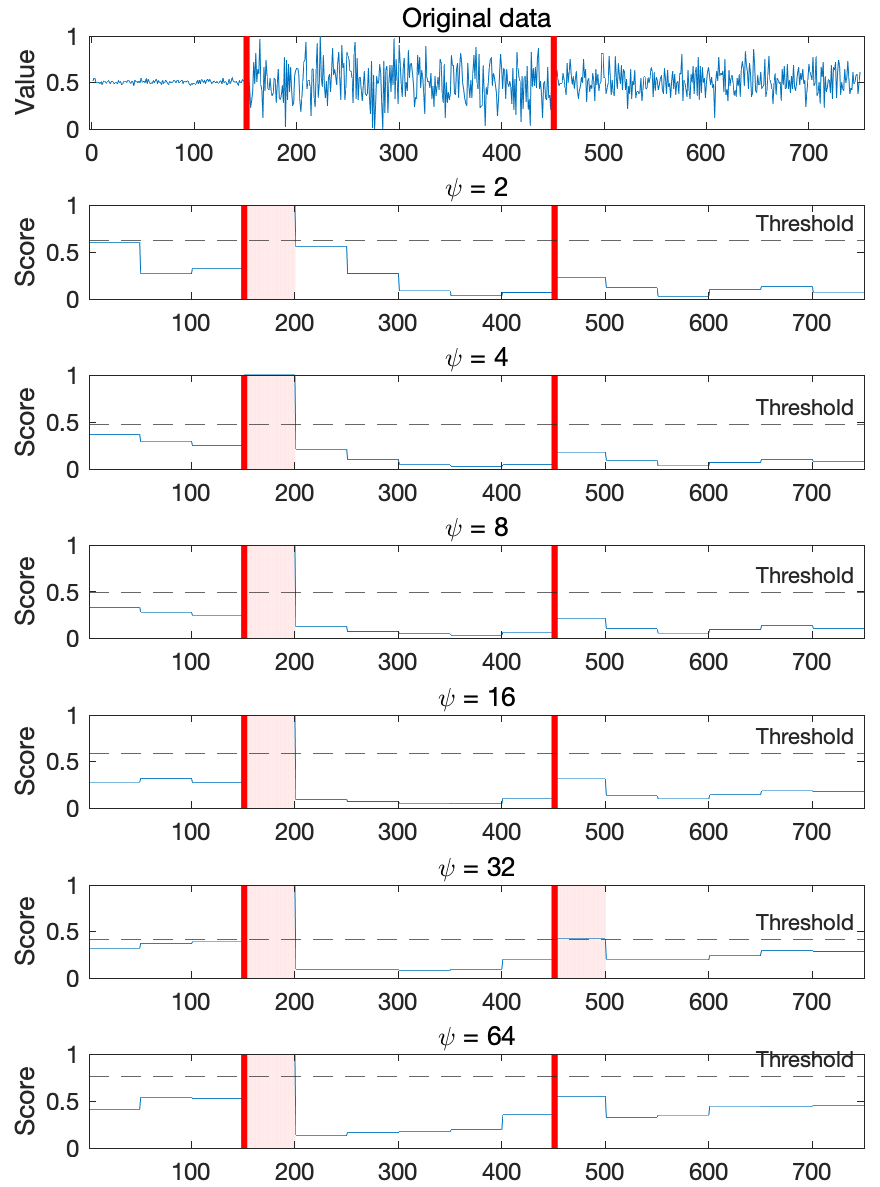}  
  \caption{S1 dataset with $w = 50$}
  \label{fig:onlinesa-syn1}
\end{subfigure}
\begin{subfigure}{.48\textwidth}
  \centering
  \includegraphics[width=0.99\linewidth]{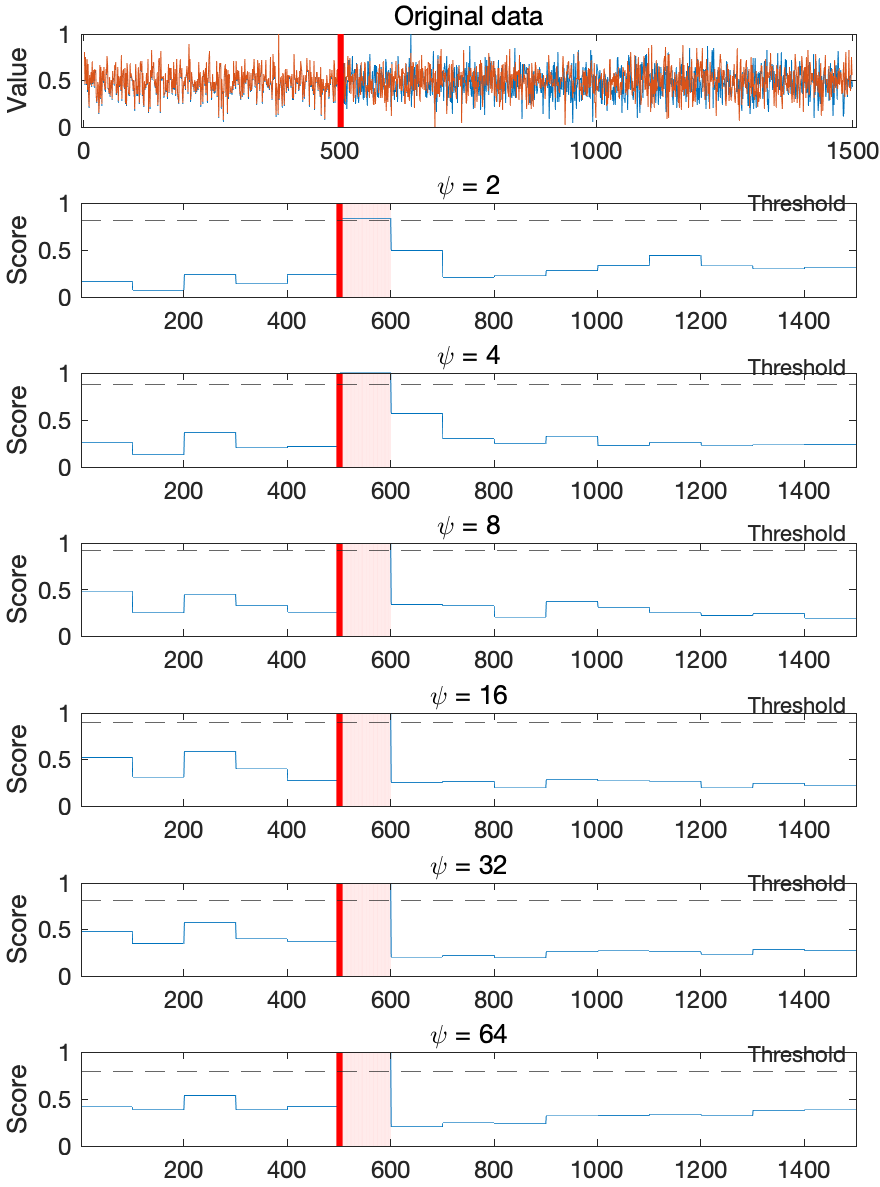}
  \caption{S2 dataset with $w = 100$}
  \label{fig:onlinesa-syn2}
\end{subfigure}
\begin{subfigure}{.48\textwidth}
  \centering
  \includegraphics[width=0.99\linewidth]{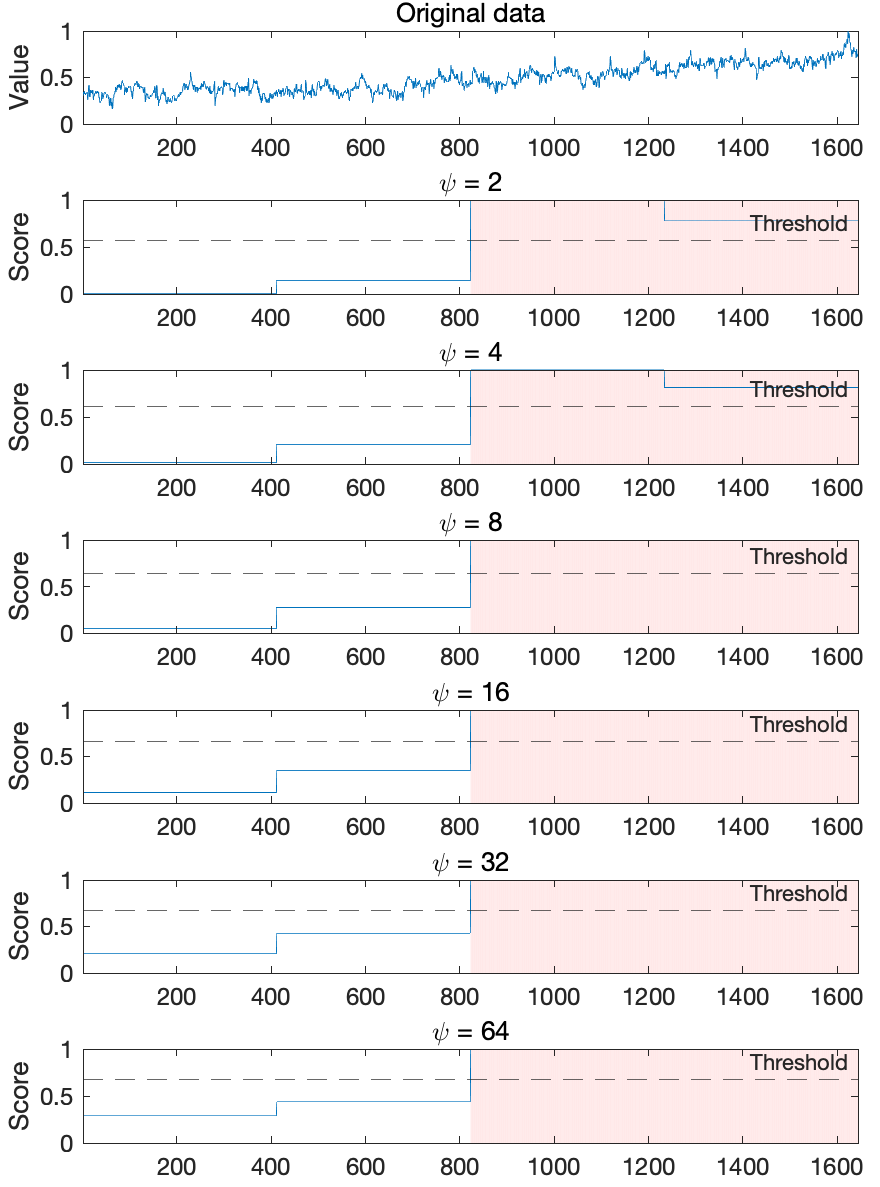}  
  \caption{Weather dataset with $w = 411$}
  \label{fig:onlinesa-weather}
\end{subfigure}
\begin{subfigure}{.48\textwidth}
  \centering
  \includegraphics[width=0.99\linewidth]{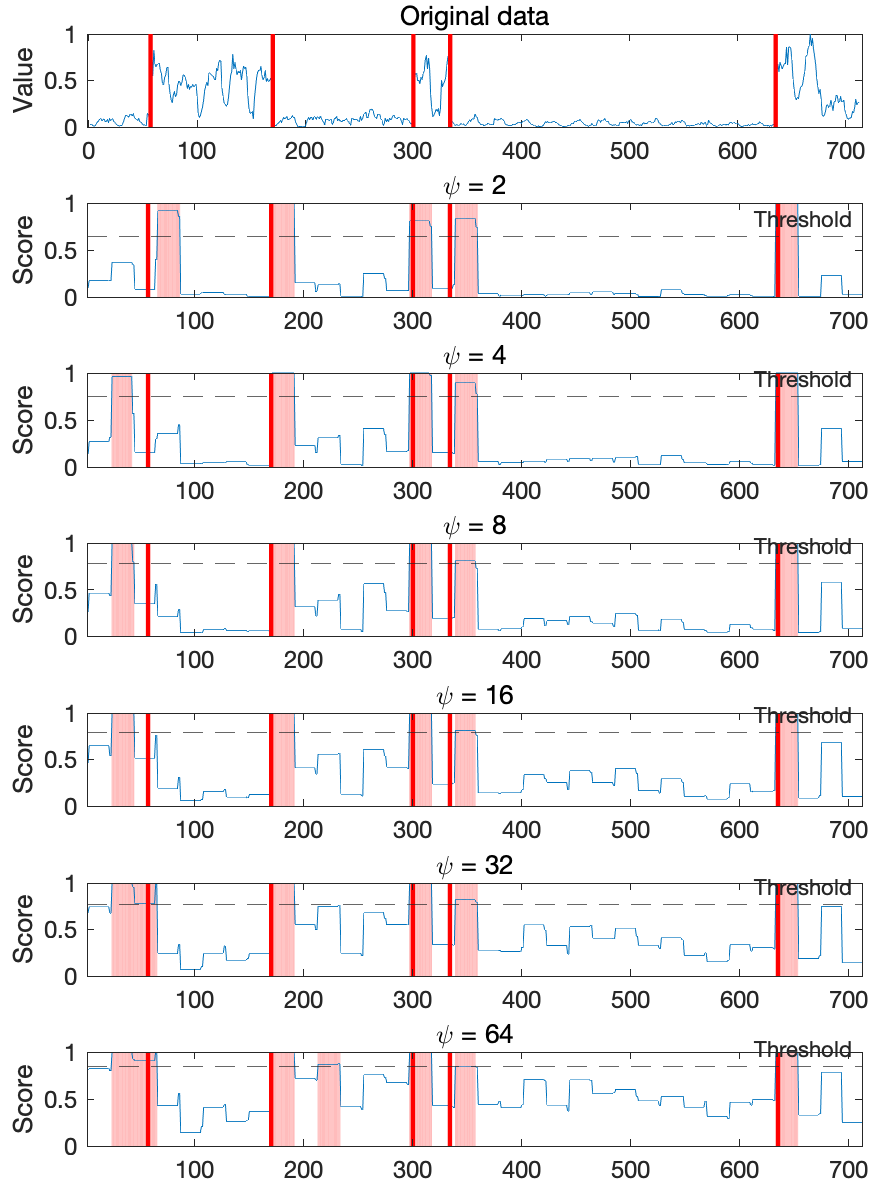}  
  \caption{Yahoo-16 dataset with $w = 21$}
  \label{fig:onlinesa-yahoo}
\end{subfigure}
\caption{Online $\texttt{iCID}$ Sensitivity analysis of $\psi$ on four datasets.}
\label{fig:online-sa}
\end{figure}

For the online \texttt{iCID}, we report the results using the $\psi^*$ searched from the first half of the data in previous experiments. It is possible to update the $\psi^*$ for each new coming interval based on the latest $k$ points, but it will cost much more running time. We report the results of online \texttt{iCID} using different $\psi$ on four datasets in Figure~\ref{fig:online-sa}. The results show that our parameter search strategy produces a similar \texttt{iCID} score to that using the best $\psi$ value.

\section{Conclusion}~\label{sec-conclusion}

In this paper, we systematically analyse three key challenges of existing change-point detection methods and propose to design a change-interval detection method to address those challenges. 

We verify that Isolation Distributional Kernel is capable of effectively measuring the dissimilarity between adjacent intervals in large data streams. Its data-dependent property enables the proposed \texttt{iCID} to detect both subtle change and obvious change-points.  \texttt{iCID} also presents promising performance with automated kernel parameter setting.

Our extensive evaluation confirms that \texttt{iCID} is effective in capturing three types of change-points in both univariate and multivariate streaming data.
In addition, \texttt{iCID} has linear runtime with regard to the size of datasets, making them suitable for detecting change intervals in large data streams. 

Therefore, \texttt{iCID} is a powerful streaming data mining tool for analysing massive streaming data, standing out for its data-dependent property and fast running speed. In the future, we will investigate the ability of \texttt{iCID} to detect drifts in the non-i.i.d. streaming data.

\section*{ACKNOWLEDGEMENTS}
This project is supported by  National Natural Science Foundation
of China (Grant No. 62076120).

\appendix

\section{CPD Algorithms Based on Different Kernels}
\label{kcpd}

\begin{figure}[!htbp]
\begin{subfigure}{\textwidth}
  \centering
  \includegraphics[width=0.8\linewidth]{figures/syn1.png} 
  \caption{S1}
  \label{fig:syn1-1}
\end{subfigure}

\begin{subfigure}{\textwidth}
  \centering
  \includegraphics[width=0.8\linewidth]{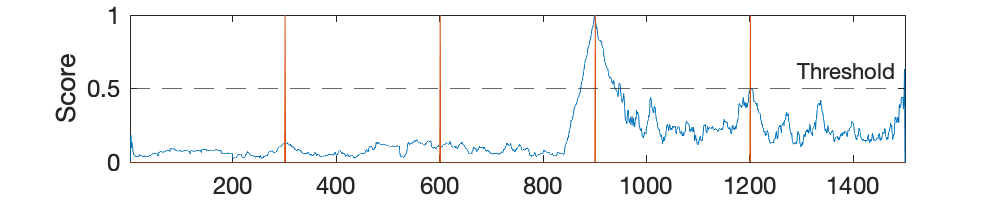}  
  \caption{Laplacian kernel}
  \label{fig:laplacian}
\end{subfigure}

\begin{subfigure}{\textwidth}
  \centering
  \includegraphics[width=0.8\linewidth]{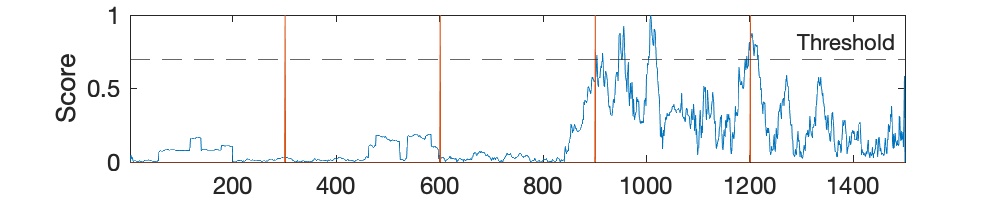}  
  \caption{Chi2 kernel}
  \label{fig:chi2}
\end{subfigure}

\begin{subfigure}{\textwidth}
  \centering
  \includegraphics[width=0.8\linewidth]{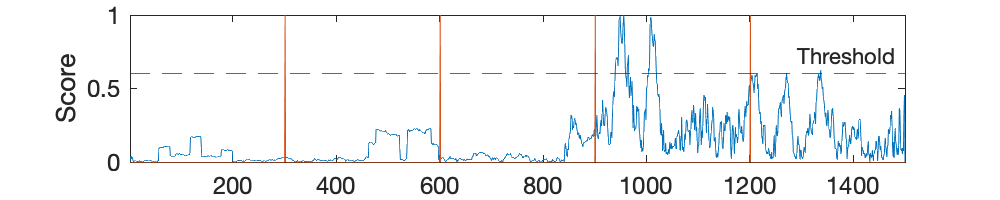}  
  \caption{Polynomial kernel}
  \label{fig:polynomial}
\end{subfigure}

\begin{subfigure}{\textwidth}
  \centering
  \includegraphics[width=0.8\linewidth]{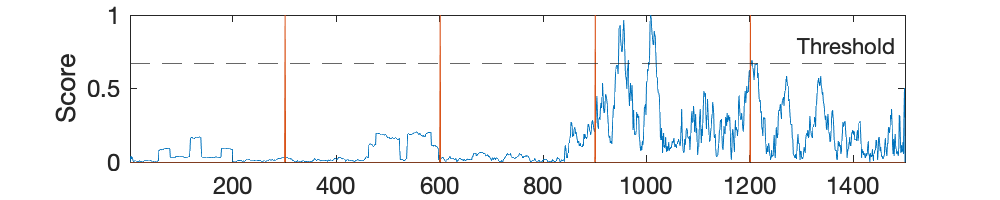}  
  \caption{Sigmoid kernel}
  \label{fig:sigmoid}
\end{subfigure}

\caption{S1 dataset: Comparison of the same proposed distributional kernel-based CID algorithm using different kernels on S1. 
}
\label{fig:kernelcpd}
\end{figure}

We compare the performance of the same CPD algorithms using 4 different kernels on the S1 dataset, including Laplacian, Chi2, Polynomial and Sigmoid kernels. Figure~\ref{fig:kernelcpd} shows the detection results indicating that none of them can reliably locate all the change-points in the dataset. This is because all these kernel methods are data-independent and have the similar issue as Gaussian kernel to identify the type I subtle-changes.

\section{Alternative Criterion For Best $\psi$ Selection}
\label{measurement}

\begin{figure}[!hbtp]
\begin{subfigure}{\textwidth}
  \centering
  \includegraphics[width=0.8\linewidth]{figures/syn1.png} 
  \caption{S1}
  \label{fig:syn1-1-1}
\end{subfigure}

\begin{subfigure}{\textwidth}
  \centering
  \includegraphics[width=0.8\linewidth]{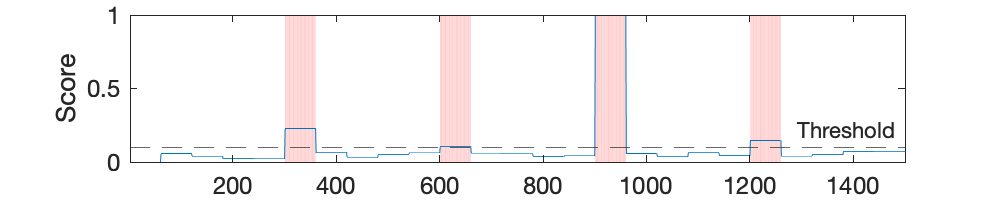}  
  \caption{Entropy}
  \label{fig:entropy}
\end{subfigure}

\begin{subfigure}{\textwidth}
  \centering
  \includegraphics[width=0.8\linewidth]{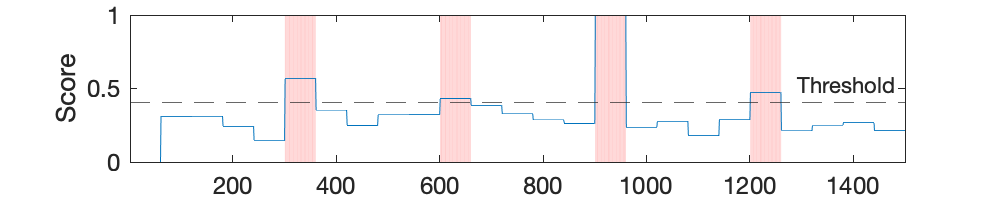}  
  \caption{Variance}
  \label{fig:variance}
\end{subfigure}

\begin{subfigure}{\textwidth}
  \centering
  \includegraphics[width=0.8\linewidth]{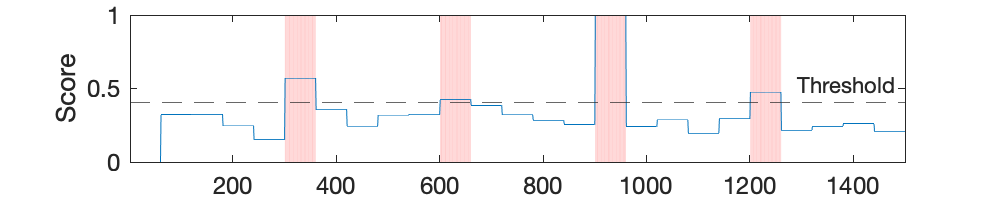}  
  \caption{Gini Coefficient}
  \label{fig:gini}
\end{subfigure}
\caption{S1 dataset: Comparison of the same proposed distributional kernel-based CID algorithm using different measurements on S1. 
}
\label{fig:measurements}
\end{figure}

We compare the experiment result of \texttt{iCID}  using three measurements for best $\psi$ selection following Equation~(\ref{eqn:psi}) on S1 dataset, including approximate entropy, variance and Gini coefficient. All these measurements can lead to the $\psi$ which can detect all change intervals on S1 dataset, which is shown in Figure~\ref{fig:measurements}.

\newpage

\bibliographystyle{theapa}
\bibliography{reference}

\end{document}